\documentclass[10pt,twocolumn,letterpaper]{article}

\usepackage[pagenumbers]{iccv} % To force page numbers, e.g. for an arXiv version

\usepackage{amssymb}
\usepackage{amsmath}
\usepackage{booktabs}
\usepackage{makecell}
\usepackage{multirow}
\usepackage{arydshln}
\usepackage{pifont}
\usepackage{parskip}
\usepackage{algorithm}
\usepackage{algpseudocode}
\usepackage{siunitx}
\usepackage[normalem]{ulem}

\definecolor{iccvblue}{rgb}{0.21,0.49,0.74}
\usepackage[pagebackref,breaklinks,colorlinks,allcolors=iccvblue]{hyperref}

\def\paperID{3360} % *** Enter the Paper ID here
\def\confName{ICCV}
\def\confYear{2025}

\title{Generative Data Augmentation for Object Point Cloud Segmentation}

\author{
Dekai Zhu$^{1,2,3}$, ~
Stefan Gavranovic$^{2}$, ~
Flavien Boussuge$^{2}$, ~
Benjamin Busam$^{1}$~and~ 
Slobodan Ilic$^{1, 2}$\\
\\[-0.5em]
$^1$ Technical University of Munich\quad
$^2$ Siemens AG\quad
$^3$ Munich Center for Machine Learning\\
}

\definecolor{dk}{rgb}{0.0,0.0,0.0}
\newcommand{\dk}[1]{\textcolor{dk}{\textnormal{{#1}}}}
\begin{document}
\maketitle
\begin{abstract}
Data augmentation is widely used to train deep learning models to address data scarcity. However, traditional data augmentation (TDA) typically relies on simple geometric transformation, such as random rotation and rescaling, resulting in minimal data diversity enrichment and limited model performance improvement. 
State-of-the-art generative models for 3D shape generation rely on the denoising diffusion probabilistic models and manage to generate realistic novel point clouds for 3D content creation and manipulation.
Nevertheless, the generated 3D shapes lack associated point-wise semantic labels, restricting their usage in enlarging the training data for point cloud segmentation tasks. 
To bridge the gap between data augmentation techniques and the advanced diffusion models, we extend the state-of-the-art 3D diffusion model, Lion, to a \textbf{part-aware generative model} that can generate high-quality point clouds conditioned on given segmentation masks. 
Leveraging the novel generative model, we introduce a \textbf{3-step generative data augmentation (GDA) pipeline} for point cloud segmentation training. 
Our GDA approach requires only a small amount of labeled samples but enriches the training data with \textbf{generated variants} and pseudo-labeled samples, which are validated by a novel \textbf{diffusion-based pseudo-label filtering} method.
Extensive experiments on two large-scale synthetic datasets and a real-world medical dataset demonstrate that our GDA method outperforms TDA approach and related semi-supervised and self-supervised methods.
\end{abstract}    
\section{Introduction}
\label{sec:intro}
\begin{figure}[t]
    \centering
    \includegraphics[width=\linewidth]{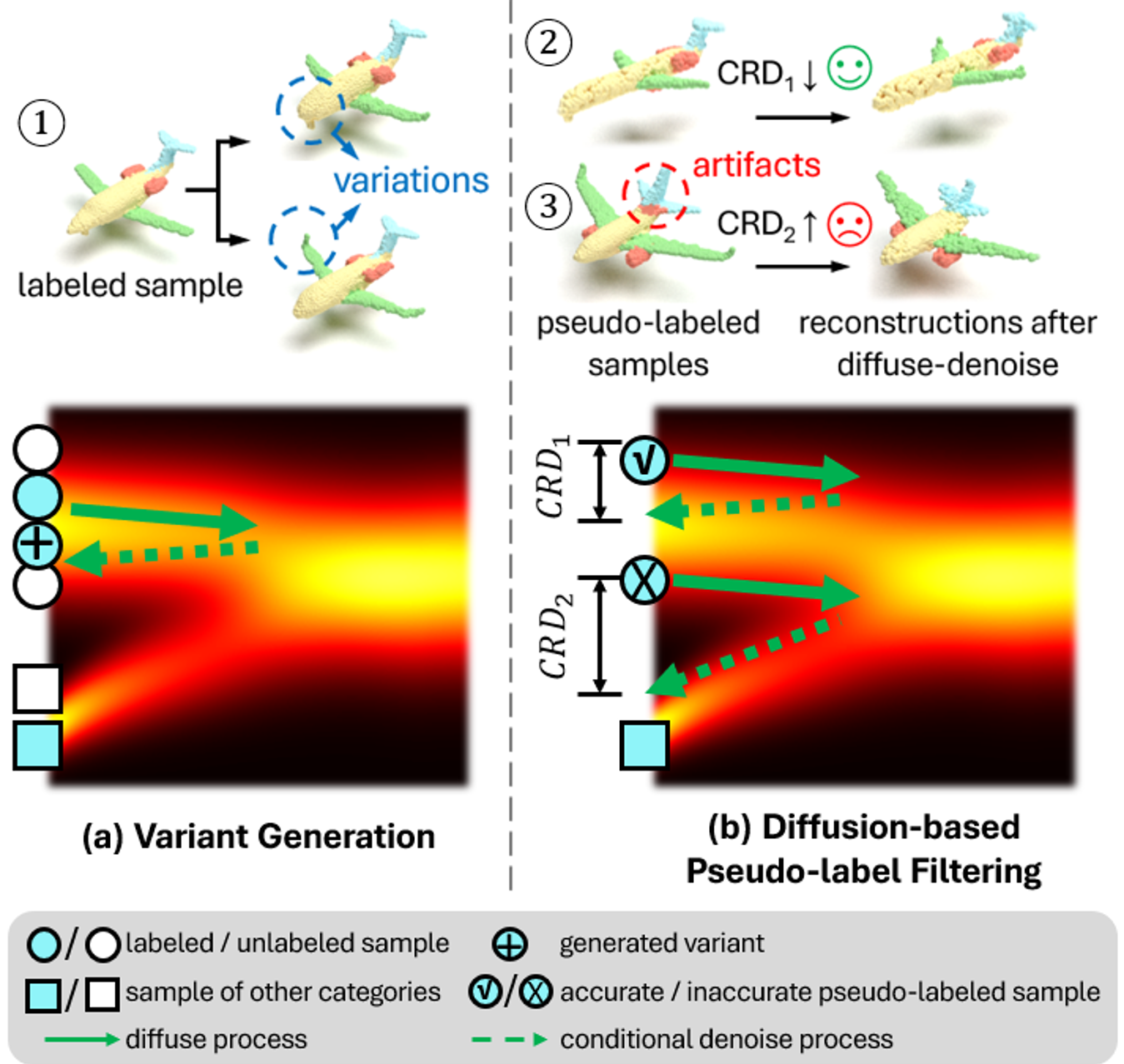}
    \caption{\dk{(a)~\textbf{Variant generation} enriches the labeled data by interpolating novel samples between the existing labeled and unlabeled samples. 
    The generated shapes exhibit diverse geometric variations. 
    (b) \textbf{Diffusion-based pseudo-label filtering} method assesses the quality of pseudo labels based on the conditional reconstruction discrepancy (CRD), calculated by the shape deviation after the conditional diffusion-denoise process guided by the pseudo label.
    The sample is closely reconstructed if the pseudo label is accurate.
    Hence, a lower CRD indicates higher pseudo-label quality, and vice versa.
    }
    }
    \label{fig:teaser}
\end{figure}
Deep learning has achieved remarkable progress across many domains~\cite{huang2025lpcg, zhu2023ipcc, zhu2024multi, zhu2025sealion, zhu2025spiral}, largely driven by the availability of large-scale annotated datasets.
However, in 3D computer vision, creating semantic segmentation annotations remains costly and time-consuming.
A straightforward way to get more annotated data is to perform traditional data augmentation (TDA), which employs basic geometric transformations, like random rescaling, rotation, flipping, jittering, etc., on already labeled data. However, this is not very useful since the variety of 3D shapes stayed small. Other approaches focused on extracting helpful information from samples with and without semantic 
segmentation annotations~\citep{jiang2021guided, hu2022sqn, cheng2021sspc, chen2019bae, di2023shapemaker, wang2020few, chen2021shape}. 
These methods can be broadly classified into semi-supervised learning~\citep{jiang2021guided, wang2020few}, weakly-supervised learning~\citep{hu2022sqn, cheng2021sspc}, and self-supervised learning~\citep{chen2019bae, chen2021shape, recon, pointcmae}. 
The semi-supervised method refers to annotating only a subset of samples, such as selecting 5\% to 10\% of available 3D objects, and training models on both labeled and unlabeled data.
Weakly-supervised methods denote those that annotate only a small number of points in each object or scene, with the most extreme case being the annotation of just one point per part, i.e., ``one part, one click".
The self-supervised method initially pretrains the segmentation model on unannotated point clouds, followed by fine-tuning on a few annotated samples. These methods have achieved superior results compared to TDA techniques, but they still have limitations, such as sensitivity to the quality of pseudo labels \citep{ssl_overview}.

In recent years, generative models represented by variational autoencoders (VAEs)~\citep{kingma2014semi}, generative adversarial networks (GANs)~\citep{goodfellow2020generative}, and denoising diffusion probabilistic models (DDPMs)~\citep{ho2020denoising} have achieved impressive performance. This provides a new perspective and can be used to bridge the gap between data annotation and data generation. This has been explored in works such as~\citep{you2023diffusion, zheng2024toward, azizi2023synthetic}. 
Experimental results in~\citep{zheng2024toward} demonstrate that the latest diffusion model~\citep{karras2022elucidating} can generate high-quality samples for data augmentation in downstream 2D classification tasks, improving the prediction accuracy. However, these works focus on 2D global classification tasks, which are more straightforward than 3D segmentation tasks. Existing 3D generative diffusion models have achieved outstanding results on point cloud-based 3D shape generation~\citep{zhou20213d, luo2021diffusion, zeng2022lion}.
Nevertheless, these works focus on generating 3D shapes without any part segmentation information, which cannot facilitate the training of 3D segmentation models.

Therefore, this work aims to generate high-quality 3D shapes conditioned on given segmentation annotations in cases where only a limited amount of labeled data is available.
\dk{We extend the state-of-the-art point cloud generation model, Lion~\citep{zeng2022lion}, on two levels. At the global encoding level, we integrate the statistical information of individual semantic parts into the global features, making it a more informative prior. 
At the local encoding level, we upgrade the original backbone PVCNN~\citep{liu2019point} to \textbf{part-aware PVCNN (p-PVCNN)}, which enables the model to learn the association between semantic part labels and point-wise geometric features, such as 3D position.}
\dk{Leveraging this part-aware generative model, we introduce a \textbf{3-step generative data augmentation (GDA) pipeline} that includes: 
\textbf{1.}~Semi-supervised training of the generative model.
\textbf{2.}~\textbf{Variant generation}, which enriches the labeled dataset by interpolating novel samples between existing labeled and unlabeled samples, as depicted in Figure~\ref{fig:teaser}~(a).
\textbf{3.}~\textbf{Diffusion-based pseudo-label filtering}, which removes inaccurate pseudo-labeled samples. We assess pseudo-label quality using \textbf{conditional reconstruction discrepancy} (\textbf{CRD}), measuring deviation after the conditional diffuse-denoise process guided by the pseudo labels, as depicted in Figure~\ref{fig:teaser}~(b).}
We conduct extensive experiments on various categories in two large-scale synthetic datasets, ShapeNetPart~\citep{Yi2016ASA} and PartNet~\citep{mo2019partnet}, as well as on a real-world 3D intracranial aneurysm dataset, IntrA~\citep{yang2020intra}. 
The results confirm the superiority of GDA over TDA and related semi-supervised~\citep{jiang2021guided} and self-supervised~\citep{recon, pointcmae} learning methods. 
We validate the feasibility of GDA on various point cloud segmentation models, including PointNet~\citep{Qi2016PointNetDL}, PointNet++~\citep{Qi2017PointNetpp}, Point Transformer~\citep{pointtransformer} and the state-of-the-art SPoTr~\citep{Park2023SelfPositioningPT}.
Additionally, we demonstrate the robustness of GDA in the challenging case where objects are arbitrarily oriented and analyze performance influencing factors of GDA, including the number of diffusion steps, the selection of hand-labeled samples, and the quality of generated labeled samples. 
In summary, the contributions of this work are the following: 

\begin{itemize}
\item A novel part-aware generative model that generates point clouds conditioned on given segmentation masks.
\item A novel 3-step GDA pipeline that generates labeled 3D shapes from a small portion of labeled 3D shapes and validates the quality of pseudo-labeled samples.
\item Extensive experiments on two large-scale datasets and a challenging real-world medical dataset show the state-of-the-art performance of our GDA method over TDA and related semi-supervised and self-supervised methods.
\item \dk{The first systematic analysis of the performance influencing factors of 3D GDA, providing valuable insights for the broader application of 3D GDA.}
\end{itemize}

\section{Related Works}

% \vspace{0.5em}
\noindent \textbf{Learning 3D segmentation from limited data.}
As mentioned in the last section, methods for learning from limited 3D data can be categorized into semi-supervised, weakly-supervised, and self-supervised methods.

In semi-supervised methods, only a small portion of samples are annotated. 
These methods leverage labeled data for supervision and use contrastive learning~\citep{xie2020pointcontrast} to learn representations from unlabeled data.
\citep{jiang2021guided} first trains a segmentation network on labeled samples, then generates pseudo labels for unlabeled data, using predictions with high confidence to guide contrastive learning.

Weakly-supervised methods provide only incomplete annotations on each sample, e.g., sparse segmentation labels.
Approaches are proposed for static objects~\citep{xu2020weakly}, indoor~\citep{hu2022sqn, liu2021one,yang2022mil} and outdoor~\citep{liu2022less} environments.
These works try to propagate labels or gradients from the labeled points to their neighboring points during training by building spatio-temporal consistencies~\citep{bastian2023segmentor}.

BAE-NET~\citep{chen2019bae} is a representative of self-supervised methods. It transforms the segmentation problem into learning a set of indicator functions for the universal parts.
\dk{ReCon \citep{recon} effectively combines contrastive learning with masked auto-encoding.
More recently, Point-CMAE \citep{pointcmae} follows this paradigm and further encourages the model to generate identical tokens when the samples are masked differently.}

% \vspace{0.5em}
\noindent \textbf{Diffusion-based 3D generation.} DPM~\citep{luo2021diffusion} and Point-Voxel Diffusion~\citep{zhou20213d} train a diffusion model directly on point clouds,
while Lion~\citep{zeng2022lion} achieves higher generation quality by utilizing a hierarchical VAE and latent diffusion models.
More recently, \citep{yu2022legonet, tang2023diffuscene, zhai2024commonscenes, zhai2024echoscene} focus on 3D scene generation. Due to higher semantics and geometry complexity, instead of directly generating a scene, these works rearrange indoor furniture into reasonable placements.

% \vspace{0.5em}
\noindent \textbf{Generative data augmentation.}
While augmentation techniques based on VAEs or GANs boost performance for 3D domain adaptation~\citep{lopez2020project, lehner20223d, lehner20243d}, these techniques do not work equally well for GDA. Empirical results~\citep{zheng2024toward} show promise using diffusion models.
Besides, \citep{you2023diffusion, azizi2023synthetic, trabucco2023effective, he2022synthetic} investigate how to use diffusion models for generative data augmentation on 2D images. 
\section{Methodology}
\begin{figure}[t]
    \centering
    \subfloat[Architecture of the part-aware generative model.
    The extended data flows for part-awareness, highlighted in \textcolor{red}{red} solid arrows, show that segmentation encoding $y$ is the input for all modules in both global and local encoding levels.
    Note that $y$ is zero padding for unlabeled samples.\\
    ]{
        \includegraphics[clip,width=0.98\columnwidth]{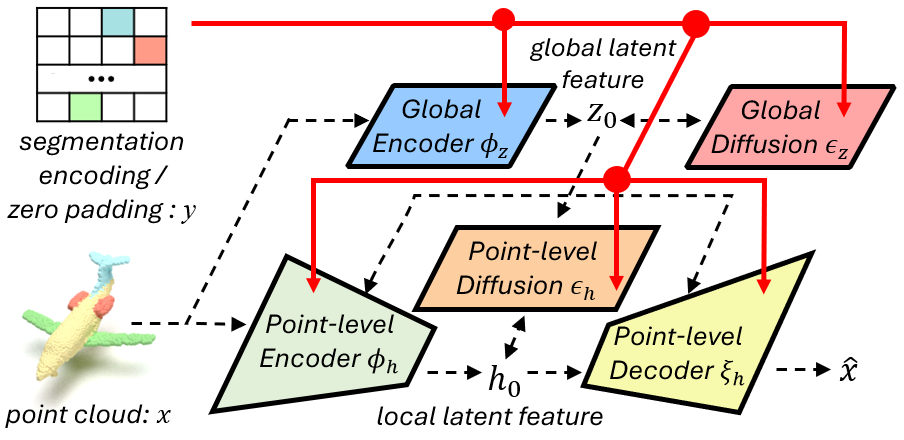}
    }
    
    \subfloat[Architecture of p-PVCNN in the point-level encoder $\phi_h$. 
    The extended modules are highlighted in \textcolor{red}{red} boxes: 1. Segmentation conditioning (SC) modules are inserted in all layers to incorporate the segmentation mask~$y$. 2. Global attention (GA) modules are used in all layers to ensure the points of small sub-parts can encode sufficient intra-part information.
    Note that the point-level decoder $\xi_h$ and diffusion $\epsilon_h$ adopt a similar architecture.]{
        \includegraphics[clip,width=0.98\columnwidth]{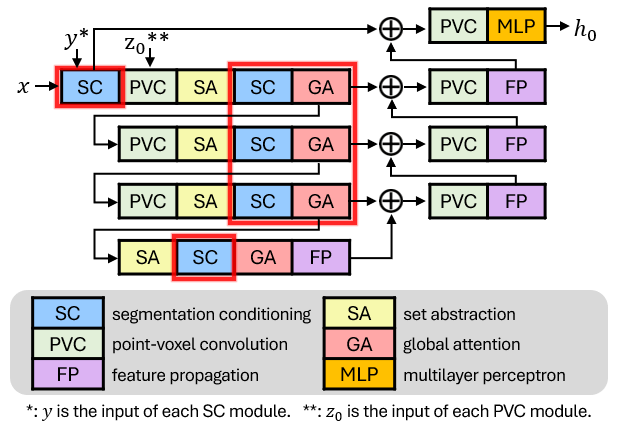}
    }
    \caption{(a) The extended Lion and (b) p-PVCNN. }
    \label{fig:pvcnn}
\end{figure}

In this section, we first introduce DDPMs~\citep{ho2020denoising} and the diffuse-denoise process.
Next, we improve Lion~\citep{zeng2022lion} at both global and local encoding levels (Figure~\ref{fig:pvcnn}) to enable it to generate labeled point clouds given segmentation masks.
Based on this model, we propose a 3-step GDA pipeline for point cloud segmentation (Figure~\ref{fig:four_step}).

\subsection{Preliminaries}
\noindent \textbf{Denoising diffusion probabilistic model.}
The diffusion model~\citep{ho2020denoising} generates data by simulating a stochastic discrete process: Given a sample $x_0 \sim q(x_0)$ from a distribution, the forward process gradually adds noise to the input to make it converge to a Gaussian distribution after $T$ steps, i.e., $q(x_T) \approx \mathcal{N} (0, I) $.
The training objective on the diffusion model $\epsilon_{\theta}$ with parameters $\theta$ is to predict the noise $\epsilon$ for the perturbed sample $x_t$:
\begin{equation}
    \mathcal{L} = \mathbb{E}_{t, x_0, \epsilon} [{|| \epsilon_{\theta} (x_t, t, a) - \epsilon || }^2_2],
\end{equation}
where $t$ is the time step uniformly sampled from $\{ 1, ..., T \}$, 
$\epsilon \sim \mathcal{N}(0, I)$ is the noise for diffusing $x_0$ to $x_t$,
and $a$ is the condition information.
During inference, the diffusion model starts from a random sample $x_T \sim \mathcal{N}(0, I)$ and denoises it iteratively until $t = 0$.
At each step, noise $\eta \sim \mathcal{N}(0, I)$ is added to the intermediate result $x_t$ to increase the randomness of the denoising process.

\noindent \textbf{Diffuse-denoise process.}
Unlike the standard inference process of DDPMs that starts from Gaussian noise and iterates for $T$ steps, the diffuse-denoise process~\citep{meng2022sdedit, zeng2022lion} diffuses an existing sample $x_0$ for $\tau$ steps ($\tau < T$) and then denoises the noisy feature $x_{\tau}$ backwards for same steps to obtain $x_{0}'$. 
During the forward diffusion process, more and more details of the input data deteriorate, while in the reverse process, besides denoising, random noise $\eta$ is added to the sample at each step to increase the variety of denoised samples.
Therefore, the diffuse-denoise process can be utilized to synthesize different variants of a given shape.

\subsection{Part-aware Generative Model}
\dk{Lion~\citep{zeng2022lion} encodes a point cloud $x$ to a global latent feature~$z_0$ and local latent features $h_0$ (referred to as ``latent points'' in~\citep{zeng2022lion}) by using a global encoder $\phi_z$ and a point-level encoder $\phi_h$ respectively.
Global and point-level diffusion modules,~$\epsilon_z$ and $\epsilon_h$, diffuse on $z_0$ and $h_0$, while the point-level decoder~$\xi_h$ reconstructs the point cloud based on these two latent features.
However, none of the modules take semantic part information into account, which hinders the generation of labeled point clouds for data augmentation in segmentation training.
To address this, we introduce the segmentation label $y$ into all modules in both global and local encoding levels, extending Lion to a part-aware generative model, as depicted in Figure~\ref{fig:pvcnn} (a).}

\noindent \textbf{Global level encoding.}
\dk{The global encoder $ \phi_z$ in Lion~\citep{zeng2022lion} utilizes the max pooling to generate the global latent feature~$z_0$, as in PointNet~\citep{Qi2016PointNetDL}, while the global diffusion $\epsilon_z$ uses a stack of ResNet~\citep{he2016deep} blocks to diffuse $z_0$.
To incorporate the segmentation encoding $y$: \textbf{1.} For $\phi_z$, we concatenate $y$ with point cloud $x$ as inputs. 
\textbf{2.} For $\epsilon_z$, we first convert $y$ into a part distribution vector $\sigma_y \in \mathbb{R}^c$ and then concatenate $\sigma_y$ with $z_t$ at step~$t$, where $c$ is the number of parts.
$\sigma_y$ represents the statistics of the number of points belonging to different parts.
This distribution varies significantly among samples within the same class.
For example, in the car class, a bus has a larger roof than most other cars, while a racing car lacks a roof.
Thus, incorporating such an informative global prior is crucial for downstream local-level encoding.
More details about the architecture the global level components are available in the supplementary materials.}

\noindent \textbf{Local level encoding.}
\dk{The point-level modules $\phi_h$, $\xi_h$, and $\epsilon_h$ in Lion utilize a tailored 4-layer PVCNN~\citep{liu2019point} as their backbones.
Each layer of PVCNN consists of point-voxel convolution (PVC) modules~\citep{liu2019point} (except the deepest layer) for local points encoding and global latent $z_0$ conditioning, along with set abstraction (SA) and feature propagation (FP) modules~\citep{Qi2017PointNetpp} for downsampling and upsampling points.
Besides, a global attention (GA) module is used in the deepest layer for self-attention encoding.
To incorporate the segmentation encoding $y$, we integrate the \textbf{segmentation conditioning module} (\textbf{SC}) before the PVC encoding modules in each layer.
In SC modules, the segmentation encoding is concatenated with the intermediate features.
This extra information brings two benefits: \textbf{1.} It reduces the uncertainty in predicting per-point geometric features, such as 3D position. \textbf{2.} It facilitates the modeling of inter-part relations, enhancing the inter-part coherence of the generated shapes.
Additionally, we observe that the number of points belonging to different parts is often unbalanced on many shapes. 
For example, the {engine} part of the {airplane} class is relatively small and thus contains fewer points than larger parts. 
As the number of points decreases in deeper layers, such small-part points tend to be underrepresented.
To mitigate this issue, we extend the use of the GA module beyond the deepest layer, applying it across \textbf{all layers}.
This ensures that points from smaller parts can encode sufficient intra-part information throughout the encoding process.
The architecture of the extended \textbf{part-aware PVCNN} (\textbf{p-PVCNN}) is illustrated in Figure~\ref{fig:pvcnn} (b).
More details about the architecture of the local level components are available in supplementary materials.}

\begin{figure*}[t]
    \centering
    \includegraphics[width=0.99\linewidth]{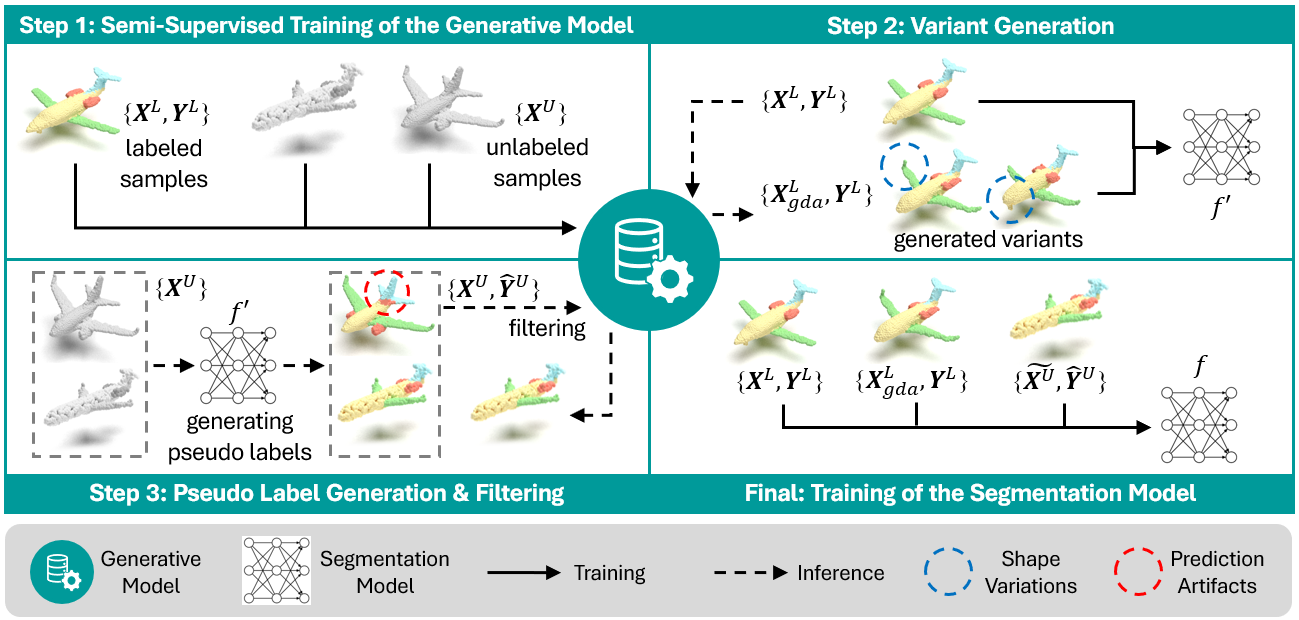}
    \caption{Pipeline of 3-step GDA. 
    \textbf{Step 1}: Using hand-labeled samples $\{ \textbf{X}^L, \textbf{Y}^L \}$ and unlabeled samples $\{ \textbf{X}^U \}$ to train the generative model in a semi-supervised approach. 
    \textbf{Step 2}: Generating variants $\textbf{X}^L_{gda}$ by running diffuse-denoise process on $\textbf{X}^L$ conditioned by $\textbf{Y}^L$, then using $\{ \textbf{X}^L, \textbf{Y}^L \}$ and $\{ \textbf{X}^L_{gda}, \textbf{Y}^L \}$ to train a temporary segmentation model $f'$.
    \textbf{Step 3}: Using $f'$ to assign pseudo labels $\hat{\textbf{Y}}^U$ to $\textbf{X}^U$, then filtering out the low-quality pseudo-labeled samples with a large \textbf{conditional reconstruction discrepancy}.
    \textbf{Final}: Using hand-labeled and generated samples from step 2 \& 3 to train the eventual segmentation model $f$.
    Note that the entire process is automatic, and the generated samples can be used for training various segmentation models.}
    \label{fig:four_step}
\end{figure*}

\subsection{Three-step GDA Pipeline}
Leveraging the part-aware generative model, we propose a 3-step GDA pipeline for the point cloud segmentation task, as illustrated in Figure~\ref{fig:four_step}.

\noindent \textbf{Notions.}
Given a training set of $N$ labeled and $M$ unlabeled point clouds of 3D objects ($N < M$), consisting of $n$ points each, the learning task is to train a model to generate novel 3D shapes, represented as point clouds with associated segmentation labels. 
Note that the number of labeled point clouds $N$ is significantly lower (around 10\%) than the overall number of input point cloud shapes $N+M$. 
For $N$ labeled point clouds, they contain the point coordinates $\textbf{X}^L = \{x^{l}_i \in \mathbb{R}^{n \times 3} | i=1,...,N \}$ and the one-hot segmentation encodings $\textbf{Y}^L = \{ y^{l}_i \in \{ 0, 1 \}^{n \times c} | i=1,...,N \}$, where $c$ is the number of parts, 
e.g. the car class in ShapeNetPart~\citep{Yi2016ASA} contains four parts: hood, roof, wheels, and car body. 
For $M$ unlabeled point clouds, they only contain the point coordinates $\textbf{X}^U = \{x^{u}_j \in \mathbb{R}^{n \times 3} | j=1,...,M \}$.
To train the generative model with the labeled and unlabeled data in a unified manner,
we use zero padding $\textbf{Y}^U = \{ y^{u}_j = \textbf{0}^{n \times c} | j=1,...,M \}$ for these unlabeled point clouds.

% \vspace{0.5em}
\noindent \textbf{Step 1: Semi-supervised training of the generative model.}
Our part-aware generative model maps the point cloud $x \in \{ \textbf{X}^L, \textbf{X}^U \}$ and associated labels or zero padding $y \in \{ \textbf{Y}^L, \textbf{Y}^U\}$ into hierarchical latent feature spaces and diffuse on these latent features, composed of a global latent $z_0 \in \mathbb{R}^{d_z}$ and a point-level latent $h_0 \in \mathbb{R}^{n \times d_h}$, where $d_z$ and $d_h$ are respectively the dimensions of these two latent spaces. 
The training of the generative model consists of two stages.
In the first training stage, we train the hierarchical VAE, consisting of the global encoder $\phi_z: \mathbb{R}^{n \times 3} \times \mathbb{R}^{n \times c} \rightarrow \mathbb{R}^{d_z}$,
point-level encoder $\phi_h: \mathbb{R}^{n \times 3} \times \mathbb{R}^{n \times c} \times \mathbb{R}^{d_z} \rightarrow \mathbb{R}^{n \times d_h}$,
and point-level decoder $\xi_h: \mathbb{R}^{n \times d_h} \times \mathbb{R}^{n \times c} \times \mathbb{R}^{d_z} \rightarrow \mathbb{R}^{n \times 3}$,
to maximize a variational lower bound on the data log-likelihood (ELBO):
\begin{equation}
\begin{aligned}
    \mathcal{L}(\phi_z, \phi_h, \xi_h) =& \mathbb{E}_{p(x), q_{\phi_z}, q_{\phi_h}} \{ \log p_{\xi_h} (x|h_0, y, z_0) \\
    -& \lambda_z D_{KL}[q_{\phi_z}(z_0|x, y)|\mathcal{N}(0, I)] \\
    -& \lambda_h D_{KL} [q_{\phi_h}(h_0|x,y,z_0) | \mathcal{N}(0, I) ] \},
\end{aligned}
\end{equation}
where $p_{\xi_h}$ is the prior for reconstruction prediction, $q_{\phi_z}$ and $q_{\phi_h}$ are the posterior distribution for sampling $z_0$ and $h_0$, and $\lambda_z$ and $\lambda_h$ are the hyperparameters for balancing Kullback-Leibler regularization and reconstruction accuracy.
In the second training stage, we train two diffusion models $\epsilon_z: \mathbb{R}^{d_z} \times \mathbb{R} \times \mathbb{R}^{n \times c} \rightarrow \mathbb{R}^{d_z}$ and $\epsilon_h: \mathbb{R}^{n \times d_h} \times \mathbb{R} \times \mathbb{R}^{n \times c \times d_z} \rightarrow \mathbb{R}^{n \times d_h}$ for $z_0$ and $h_0$, respectively by minimizing the objectives:
\begin{equation}
    \mathcal{L}(\epsilon_z) = \mathbb{E}_{t, z_0, \epsilon} [{|| \epsilon_{z} (z_t, t, y) - \epsilon || }^2_2],
\end{equation}
and 
\begin{equation}
    \mathcal{L}(\epsilon_h) = \mathbb{E}_{t, h_0, \epsilon} [{|| \epsilon_{h} (h_t, t, [y, z_0]) - \epsilon || }^2_2],
\end{equation}
where $z_0 = \phi_z (x, y)$, 
$h_0 = \phi_h (x, y, z_0)$, 
$x \in \{ \textbf{X}^L, \textbf{X}^U \}$, 
$y \in \{ \textbf{Y}^L, \textbf{Y}^U \}$, 
$\epsilon \sim \mathcal{N}(0, I)$,
and $t$ is the time step uniformly sampled from $\{ 1, ..., T \}$.
The models $\epsilon_z$ and $\epsilon_h$ with parameters $z$ and $h$ are conditioned by $y$ and $[y,z_0]$ respectively.
The generative model is trained in this semi-supervised approach to learn the reconstruction of the point clouds with or without the segmentation labels.

\noindent \textbf{Step 2: Variant generation.} 
Once the generative model is trained, it can be used to run a $\tau$-step diffuse-denoise process on $\textbf{X}^L$. For each labeled sample $x^{l}_i$, we can obtain a set of variants $\textbf{X}^{l}_{i} = \{ x^{l}_{i, \tau} \in \mathbb{R}^{n \times 3} | 0<\tau \! < \! T \} $ conditioned by the same segmentation labels $y^{l}_i$.
Compared with $x^{l}_i$, the generated variants $x^{l}_{i, \tau}$ contain some local deformations that are learned from both labeled and unlabeled data. 
\dk{Thus, this process essentially performs interpolation between the existing labeled and unlabeled samples, while ensuring that the generated samples maintain a reasonable shape.
An intuitive illustration is presented in Figure~\ref{fig:teaser}~(a), where the generated shapes exhibit diverse and reasonable variations.
Through this process, segmentation annotations are properly transferred to the newly generated samples.}
With the original hand-labeled data $\textbf{X}^L$ and the novel generated variants $\textbf{X}^L_{gda} = \{ \textbf{X}^{l}_{i} | i=1,...,N \} $, we can train a temporary segmentation neural network $f': \mathbb{R}^{n \times 3} \rightarrow \{ 0, 1 \}^{n \times c} $ for predicting pseudo labels.

% \vspace{0.5em}
\noindent \textbf{Step 3: Pseudo label generation \& filtering.} 
Using $f'$, we can generate pseudo labels $\hat{\textbf{Y}}^U = \{ \hat{y}^{u}_j \in \{ 0, 1 \}^{n \times c} | j=1,...,M \} $ on unlabeled samples $\textbf{X}^U$.
The pseudo labels inevitably contain some artifacts that can degrade the training of segmentation models.
\dk{
Instead of visually inspecting the pseudo-labeled samples, we propose a novel \textbf{Diffusion-based pseudo-label filtering} method to evaluate the quality of pseudo labels:
\textbf{1. }The generative model performs a $\tau'$-step diffuse-denoise process $(0<\tau'<T)$ on each unlabeled sample $x^u_j$, \textbf{conditioned} by the their pseudo label $\hat{y}^u_j$. 
During this process, $x^u_j$ is perturbed and then reconstructed to $\hat{x}^u_j$.
Notably, the noise $\eta$ is deprecated during denoising to reduce the randomness.
\textbf{2. }We compute the deviation between $x^u_j$ and $\hat{x}^u_j$ to evaluate the quality of $\hat{y}^u_j$.
If $\hat{y}^u_j$ is accurate, the sample should be well reconstructed after this process. 
Otherwise, significant deviation occurs as $\hat{x}^u_j$ adapts to the inaccurate pseudo label. 
We define this deviation as \textbf{conditional reconstruction discrepancy} (\textbf{CRD}), 
where a large CRD indicates a low-quality pseudo label, and vice versa. 
An intuitive comparison is presented in Figure~\ref{fig:teaser} (b). 
The airplane \ding{173}, with accurate pseudo labels, is closely reconstructed, whereas the substantial change in the engine (\textcolor[RGB]{240, 67, 50}{red}) part's volume in \ding{174} results in a large CRD, indicating the inaccuracy of its pseudo labels.
CRD is calculated based on the voxelized IoU between $x^u_j$ and $\hat{x}^u_j$.
Specifically, we compute the IoU for each part separately and then average the results to mIoU to ensure that smaller parts are also sufficiently considered.
\textbf{3.} The pseudo-labeled samples with mIoU between $x^u_j$ and $\hat{x}^u_j$ below the threshold $\delta$ are filtered out.}

% \vspace{0.5em}
\noindent \textbf{Final: Training of the segmentation model.} 
After 3-step GDA, the original training set only containing the hand-labeled data $\textbf{X}^L$ is enlarged with the generated variants $\textbf{X}^L_{gda} = \{\textbf{X}^{l}_{i} | i=1,...,N \}$ from step 2 and the filtered pseudo-labeled samples $\tilde{\textbf{{X}}^U} \subseteq \textbf{{X}}^U$ from step 3. 
Eventually, all these labeled samples are used to train the final segmentation network $f: \mathbb{R}^{n \times 3} \rightarrow \{ 0, 1 \}^{n \times c}$.

\section{Experiments}
\subsection{Experimental Settings} \label{sec:exp_setting}

\noindent \textbf{Datasets.}
We conduct experiments on two large-scale synthetic datasets, ShapeNetPart~\citep{Yi2016ASA} and PartNet~\citep{mo2019partnet}, and a real-world dataset, IntrA~\citep{yang2020intra}, which contains 3d intracranial aneurysm point clouds reconstructed from MRI.
We follow the official train/val/test splits of ShapeNetPart and PartNet.
From ShapeNetPart, we use airplane, car, lamp, motorbike, and pistol categories, and from PartNet, we use bed, bottle, chair, faucet, and table categories.
During training, 10\% of the samples in each category are provided with segmentation labels, while the remaining are unlabeled.
IntrA dataset contains 116 labeled and 215 unlabeled aneurysm segments. We use 24 labeled samples and 215 unlabeled samples for training, while the remaining 92 labeled samples are reserved for testing.

\begin{table}[t]\small
\centering
% \begin{center}
\renewcommand\arraystretch{1.1}
\begin{tabular}{l @{\hskip 0.0cm} c @{\hskip 0.25cm} c @{\hskip 0.4cm} c @{\hskip 0.15cm} c @{\hskip 0.2cm} c }
\toprule
Method        & Airplane & Car   & Lamp  & Motorbike & Pistol \\ \hline
FS              & 80.40    & 76.34 & 72.04 & 67.43     & 82.52  \\ \hdashline[1pt/2pt]
w/o aug.        & 76.46    & 70.09 & 66.04 & 56.10     & 77.66  \\ 
only TDA             & 76.84    & 71.48 & 66.71 & 59.03     & 78.91  \\ 
CL {\citep{jiang2021guided}}              & 76.49    & 71.69 & 66.78 & 59.64     & 79.10  \\ 
ReCon \citep{recon} & 77.12 & \underline{74.51} & 66.89 & \underline{65.45} & 80.93\\
Point-CMAE \citep{pointcmae} & 77.46 & 74.45 & \underline{68.23} & \textbf{65.57} & \underline{81.02} \\
GDA {\fontsize{8}{8}\selectfont (VG)}        & \underline{77.96}    & 72.90 & 66.79 & 61.44     & 80.01  \\ 
GDA {\fontsize{8}{8}\selectfont (VG+FP)}  & \textbf{78.32}    & \textbf{74.89} & \textbf{71.22} & 64.67     & \textbf{81.35}  \\ \bottomrule
\end{tabular}
\caption{Evaluation on ShapeNetPart~\citep{Yi2016ASA}. 
All mIoU scores are given in percentage (\%). 
The \textbf{best} and \underline{second-best} scores for each category are highlighted in \textbf{bold} and \underline{underlined}, respectively.}
\label{tab:shapenet}
% \end{center}
\end{table}
\begin{table}[t]\small
\centering
\renewcommand\arraystretch{1.1}
\begin{tabular}{l @{\hskip 0.2cm} c @{\hskip 0.4cm} c @{\hskip 0.4cm} c @{\hskip 0.4cm} c @{\hskip 0.4cm} c }
\toprule
Method        & Bed   & Bottle & Chair & Faucet & Table \\ \hline
FS              & 38.29 & 52.91  & 60.66 & 53.19  & 58.07 \\ \hdashline[1pt/2pt]
w/o aug.        & 27.36 & 40.99  & 51.69 & 40.01  & 46.28 \\ 
only TDA             & 28.19 & 43.06  & 53.25 & 46.20  & 47.89 \\ 
CL {\fontsize{8}{8}\selectfont \citep{jiang2021guided}}              & 29.00 & 47.49  & 53.44 & 47.19  & 50.36 \\ 
ReCon \citep{recon} & 30.88 & 49.37 & 58.65 & 47.59 & \underline{51.02} \\
Point-CMAE \citep{pointcmae} & \underline{31.53} & 49.54 & \textbf{59.23} & 47.26 & \textbf{51.17}\\
GDA {\fontsize{8}{8}\selectfont (VG)}        & 30.69 & \underline{50.14}  & 56.40 & \underline{49.25}  & 50.22 \\ 
GDA {\fontsize{8}{8}\selectfont (VG+FP)} & \textbf{32.96} & \textbf{53.21}  & \underline{58.72} & \textbf{50.31}  & 50.42 \\ \bottomrule
\end{tabular}
\caption{Evaluation on PartNet~\citep{mo2019partnet}. 
All mIoU scores are given in percentage (\%). 
The \textbf{best} and \underline{second-best} scores for each category are highlighted in \textbf{bold} and \underline{underlined}, respectively.}
\label{tab:partnet}
\end{table}

\noindent \textbf{Segmentation network.}
We primarily use PointNet~\citep{Qi2016PointNetDL} as the segmentation model. 
In the ablation studies, we additionally test PointNet++~\citep{Qi2017PointNetpp}, Point Transformer~\citep{pointtransformer}, and SPoTr~\citep{Park2023SelfPositioningPT}, which is a state-of-the-art and open-source model on the ShapeNetPart segmentation benchmark~\citep{paperswithcode_shapenet_part}. 

\begin{figure}[t]
    % \centering
    \begin{center}
    \includegraphics[width=0.95\linewidth]{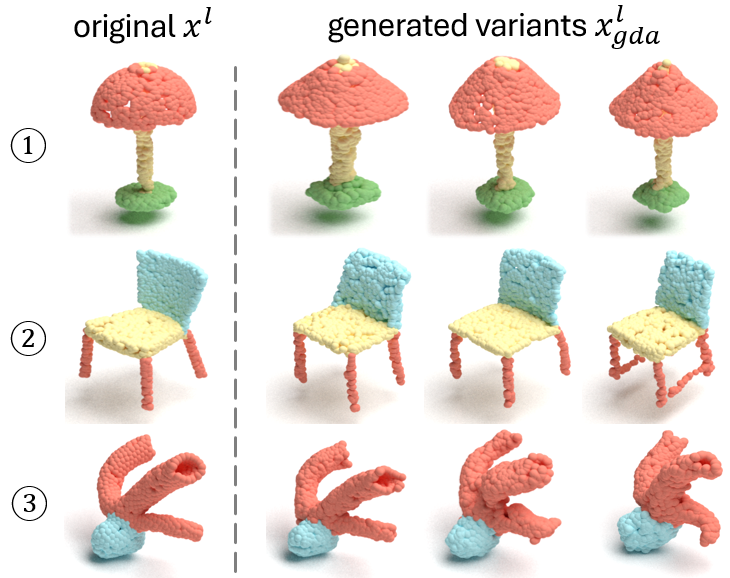}
    \caption{Illustration of hand-labeled point clouds $x^l$ (the leftmost column) and their corresponding generated variants $x^l_{gda}$ (the right three columns). \ding{172} A lamp from ShapeNetPart~\citep{Yi2016ASA}. \ding{173} A chair from PartNet~\citep{mo2019partnet}. \ding{174} An aneurysm segment from IntrA~\citep{yang2020intra} (\textcolor[RGB]{240, 67, 50}{red}: vessels, \textcolor[RGB]{125, 200, 255}{blue}: aneurysm).}
    \label{fig:gda_variants}
    \end{center}
\end{figure}

% \vspace{0.5em}
\noindent \textbf{Methods.}
We evaluate the performance of the following methods in the experiments:\\
\textbf{1.}~Full supervision (\textbf{FS}): using the whole training set;\\
\textbf{2.}~Without data augmentation (\textbf{w/o aug.}): using 10\% labeled samples without any data augmentation method;\\
\textbf{3.}~\textbf{Only TDA}: using 10\% labeled samples and TDA methods, including random rescaling (0.8, 1.2), random transfer (-0.1, 0.1), jittering (-0.005, 0.005), and random flipping;\\
\textbf{4.}~Contrastive learning (\textbf{CL}): using 10\% labeled samples and 90\% unlabeled samples with the method in~\citep{jiang2021guided};\\
\dk{\textbf{5.}~\textbf{ReCon}: pre-training on the whole training set without labels and fine-tuning on 10\% labeled samples with the self-supervised method proposed in~\citep{recon}; \\
\textbf{6.}~\textbf{Point-CMAE}: pre-training on the whole training set without labels and fine-tuning on 10\% labeled samples with the self-supervised method proposed in~\citep{pointcmae};} \\
\textbf{7.}~\textbf{GDA} based on variant generation (\textbf{VG}): using 10\% labeled data and the generated variants $\textbf{X}^L_{gda}$;\\
\textbf{8.}~\textbf{GDA} based on variant generation and filtered pseudo labels~(\textbf{VG+FP}): using 10\% labeled samples, $\textbf{X}^L_{gda}$, and the filtered pseudo-labeled samples~$\tilde{\textbf{X}^U}$.
Note that TDA is used in \textbf{all} methods above except method 2.

\begin{figure}[t]
    \centering
    \includegraphics[width=0.98\linewidth]{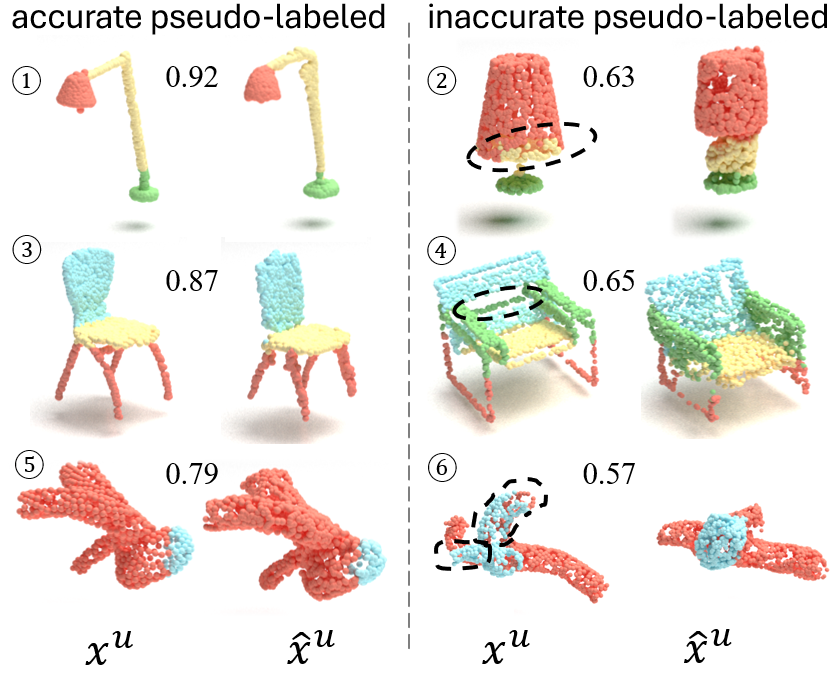}
    \caption{Illustration of the pseudo-labeled samples $x^u$ and the reconstructed samples $\hat{x}^u$ after the diffuse-denoise process conditioned on pseudo labels $\hat{y}^u$.
    Samples $x^u$ with accurate $\hat{y}^u$ are closely reconstructed, whereas those with inaccurate $\hat{y}^u$ undergo substantial shape deviation, especially in the misclassified parts, highlighted by \dotuline{black dashed circles}. 
    The values represent the voxelized IoU between $x^u$ and $\hat{x}^u$. 
    A \textbf{higher} IoU indicates a \textbf{lower} conditional reconstruction discrepancy (CRD) and \textbf{higher}-quality pseudo labels, and vice versa. }
    \label{fig:refinement}
\end{figure}

\noindent \textbf{Experimental Details.} The training of our generative model includes two stages. 
The VAE modules are trained for 8k epochs in the first stage, and the latent diffusion modules are trained for 24k epochs in the second stage. An Adam optimizer with a learning rate of $\text{1e-3}$ is utilized in these two stages.
The parameter sizes of the VAE and diffusion modules are 22.3M and 88.9M, respectively.
We set the mIoU threshold~$\delta$ to 0.7 for pseudo-label filtering.
For the segmentation models, we train them~\citep{Qi2016PointNetDL, Qi2017PointNetpp, pointtransformer, Park2023SelfPositioningPT} following the official default settings.
We conduct the experiments using an NVIDIA RTX 3090 GPU with 24GB of VRAM.

\subsection{Main Results}
We report the experimental results of mIoU scores on ShapeNetPart~\citep{Yi2016ASA}, PartNet~\citep{mo2019partnet}, and IntrA~\citep{yang2020intra} in Table~\ref{tab:shapenet}, \ref{tab:partnet}, and~\ref{tab:intra} respectively.
Across all these datasets, GDA with variant generation and filtered pseudo labels (VG+FP) achieves the best performance in most categories. 
On average, GDA outperforms methods using only TDA and Point-CMAE~\citep{pointcmae} by \textbf{3.50\%} and \textbf{0.74\%} on ShapeNetPart, \textbf{5.41\%} and \textbf{1.38\%} on PartNet, \textbf{6.38\%} and \textbf{1.46\%} on IntrA dataset.
Notably, PointNet trained with 10\% hand-labeled data and GDA even surpasses the model trained with full supervision on the {bottle} class in PartNet.
To prove the stability of GDA despite the stochasticity of DDPM, we repeat the experiment on car class {10 times}. The results of using only TDA and GDA are \textbf{71.49±0.27} and \textbf{74.92±0.46}.

Figure~\ref{fig:gda_variants} presents generated variants $x^l_{gda}$ alongside their corresponding original samples $x^l$.
The generated variants exhibit realistic and diverse local deformations, such as variations in lamp shades~(\ding{172}), chair legs~(\ding{173}), and healthy vessels~(\ding{174}), enhancing the data diversity beyond simple geometric transformations.
Figure~\ref{fig:refinement} illustrates the deviation in samples~$x^u$ after the diffuse-denoise process conditioned by accurate or inaccurate pseudo labels~$\hat{y}^u$.
Samples~(\ding{172}, \ding{174}, \ding{176}) with accurate pseudo labels are closely reconstructed with high voxelized IoU scores between $x^u$ and $\hat{x}^u$, indicating a low CRD and the high quality of $\hat{y}^u$.
In contrast, parts of samples~(\ding{173}, \ding{175}, \ding{177}) that are misclassified undergo substantial deviation after this process, with points being repositioned according to the corresponding prediction. 
This results in the low voxelized IoU between $x^u$ and $\hat{x}^u$, indicating a high CRD and the low quality of $\hat{y}^u$.
More examples of generated variants and deviations in reconstructed samples are provided in supplementary materials.

\begin{table}[t]\small
\begin{minipage}[t!]{.52\linewidth}
\centering
\includegraphics[width=0.95\linewidth]{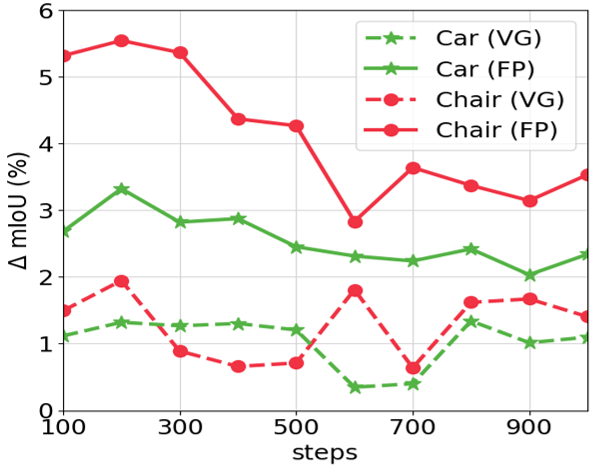}
\captionof{figure}{Impacts of the diffusion-step number on \dotuline{variant generation} and \underline{pseudo-label filtering}. The y-axis represents the mIoU improvement over TDA.}
% (\hdashrule[0.5ex]{0.5cm}{0.5pt}{0.6mm})
\label{fig:dd_step}
\end{minipage}
\hfill
\begin{minipage}[t!]{.43\linewidth}
\centering
\renewcommand\arraystretch{1.1}
\scalebox{0.9}{
\begin{tabular}[t]{ l @{\hskip -0.2cm} r }
% \begin{tabular}{l r}
\toprule
Method        & Aneurysm \\ \hline
w/o aug.        & 37.49    \\ 
only TDA        & 42.29    \\ 
CL {\fontsize{8}{8}\selectfont \citep{jiang2021guided}}             & 43.63    \\ 
ReCon \citep{recon} & 47.37 \\
Point-CMAE \citep{pointcmae} & 47.21 \\
% GDA {\fontsize{8}{8}\selectfont (LT)}        & 48.14    \\ 
GDA {\fontsize{8}{8}\selectfont (VG)} & 46.85 \\
GDA {\fontsize{8}{8}\selectfont (VG+FP)} & \textbf{48.67}    \\ \bottomrule
\end{tabular}}
\caption{Evaluation on IntrA~\citep{yang2020intra} (no `FS' since all labeled samples are used for training/test).}
\label{tab:intra}  
\end{minipage}
% \hfill
\end{table}

\subsection{Ablation Studies}

\noindent \textbf{Impacts of Diffusion-based pseudo-label filtering.}
We use the generative model as a filter to remove samples with inaccurate pseudo labels. 
In this experiment, we compare our approach with other filtering methods, including: 1.~Using all pseudo labels without any filtering, and 2.~PseudoAugment~\citep{leng2022pseudoaugment}, which filters out pseudo labels with low confidence scores. 
We conduct the experiments on the car class. 
The results of the above two methods are \textbf{73.13} and \textbf{73.46}, respectively, both of which are lower than the result of GDA (\textbf{74.89}). 

\begin{table}[t]\small
\centering
\renewcommand\arraystretch{1.2}
\scalebox{0.9}{
\begin{tabular}{l @{\hskip 0.cm} c @{\hskip 0.0cm} c @{\hskip 0.2cm} c @{\hskip 0.35cm} c @{\hskip 0.15cm} c @{\hskip 0.15cm} c }
\toprule
Backbone & Method        & Airplane & Car   & Lamp  & Motorbike & Pistol \\ \hline
PointNet++ & only TDA        &  78.10 & 71.52 & 72.64 & 57.88 & 73.24 \\ % \hline
~~~~~\citep{Qi2017PointNetpp} & GDA {\fontsize{8}{8}\selectfont (VG+FP)} & 79.92 & 74.66 & 81.15 & \textbf{65.27} & 79.53  \\ \hdashline[1pt/2pt]
\multirow{2}{*}{~~PT \citep{pointtransformer}} & only TDA        &  79.15 & 71.63 & 75.67 & 53.01 & 75.44 \\ % \hline
& GDA {\fontsize{8}{8}\selectfont (VG+FP)} & 81.06 & 75.25 & \textbf{81.42} & 64.73 & 81.07 \\ \hdashline[1pt/2pt]
\multirow{2}{*}{SPoTr \citep{Park2023SelfPositioningPT}} & only TDA        & 79.36    & 71.00 & 80.78 & 46.25     & 75.71  \\ % \hline
& GDA {\fontsize{8}{8}\selectfont (VG+FP)} & \textbf{81.25}    & \textbf{75.73} & 81.31 & 65.04     & \textbf{81.99}  \\ \bottomrule
\end{tabular}}
\caption{GDA for PointNet++ \citep{Qi2017PointNetpp}, Point Transformer \citep{pointtransformer}, and SPoTr \citep{Park2023SelfPositioningPT} on ShapeNetPart~\citep{Yi2016ASA}.}
\label{tab:spotr}
\end{table}

\noindent \textbf{Impacts of the number of diffusion steps.}
In both variant generation and pseudo-label filtering, the generative model diffuses for $\tau$ and $\tau'$ steps, respectively. 
During this process, increasing the number of diffusion steps enhances the diversity of regenerated samples. However, excessive diffusion steps may degrade shape quality, particularly if the generative model is not sufficiently well-trained.
To analyze this effect, we conduct experiments on the car and chair categories, varying $\tau$ from 100 to 1k steps in increments of 100. As shown in Figure~\ref{fig:dd_step}, the variants generated with different diffusion steps consistently contribute to downstream training. Based on this observation, we incorporate samples generated across all diffusion steps in all experiments.
For $\tau'$ varying from 100 to 1k steps in increments of 100, both categories achieve the best performance with 200 steps. Therefore, we adopt this setting in other experiments in this work.

\noindent \textbf{GDA for various segmentation models.}
In this experiment, we test the performance of GDA using PointNet++~\citep{Qi2017PointNetpp}, Point Transformer~\citep{pointtransformer}, and SPoTr~\citep{Park2023SelfPositioningPT} on ShapeNetPart.
The experimental results listed in Table~\ref{tab:spotr} show that GDA steadily enhances the performance of these models, including the SOTA~\citep{paperswithcode_shapenet_part} model, SPoTr~\citep{Park2023SelfPositioningPT}.

\noindent \textbf{Objects in arbitrary orientations.}
Objects in ShapeNetPart are presented in canonical poses, while maintaining the pose consistency of 3D objects in real-world applications is challenging.
To test the robustness of our GDA method against arbitrarily oriented objects, we randomly rotate the objects in ShapeNetPart~\citep{Yi2016ASA} and repeat the experiments. 
The results listed in Table~\ref{tab:shapenet_random_pose} show that GDA exhibits a larger advantage over TDA and Point-CMAE~\citep{pointcmae} in this challenging case, outperforming them by \textbf{20.76\%} and \textbf{1.94\%} respectively on average across categories.

\begin{table}[t]\small
\centering
\renewcommand\arraystretch{1.1}
\begin{tabular}{l @{\hskip 0.0cm} c @{\hskip 0.25cm} c @{\hskip 0.35cm} c @{\hskip 0.2cm} c @{\hskip 0.2cm} c}
\toprule
Method        & Airplane & Car   & Lamp  & Motorbike & Pistol \\ \hline
FS              & 71.98    & 72.71 & 57.53 & 60.59     & 76.99  \\ \hdashline[1pt/2pt]
w/o aug.        & 20.49    & 17.84 & 22.35 & 16.53     & 31.54  \\ 
only TDA        & 61.12    & 31.09 & 37.56 & 48.38     & 43.82  \\ 
Point-CMAE \citep{pointcmae} & 68.23 & 62.97 & 55.49 & \textbf{59.37} & 70.02 \\
GDA {\fontsize{8}{8}\selectfont (VG+FP)} & \textbf{70.47}    & \textbf{67.43} & \textbf{57.76} & 58.14     & \textbf{71.98}  \\ \bottomrule
\end{tabular}
\caption{Evaluation on ShapeNetPart~\citep{Yi2016ASA} (arbitrary orientation).}
\label{tab:shapenet_random_pose}
\end{table}

\noindent \textbf{Quality of hand-labeled and generated data.}
We have observed that some hand-labeled samples from ShapeNetPart~\citep{Yi2016ASA} contain labeling errors.
Therefore, in the experiments presented above, we inspect the quality of labels and exclude the problematic samples when selecting the hand-labeled samples $\textbf{X}^L$ for training.
To evaluate the robustness of GDA against label errors, we re-train a generative model on arbitrary-oriented cars while the hand-labeled samples (10\%) are randomly selected.
The results listed in Table~\ref{tab:label_quality} show that while GDA still outperforms TDA, it experiences a more substantial performance drop due to the presence of label errors. 
More discussions on the impacts of low-quality labeled samples on TDA and GDA are available in the supplementary materials.

\begin{table}[t!]\small
\renewcommand\arraystretch{1.1}
  \begin{minipage}[t]{.48\linewidth}
    \centering
    \begin{tabular}{l @{\hskip 6pt} r}
        \toprule
        Selection & mIoU \\ \hline
        only TDA (C)   & 31.09 \\ 
        only TDA (w/o C)  & 27.53 \\ \hdashline[1pt/2pt]
        GDA (C)   & 67.43 \\
        GDA (w/o C)  & 59.27 \\ \bottomrule
    \end{tabular}
    \caption{Impact of the quality of original labeled data (C: labeled data is selected with check).}
    \label{tab:label_quality}
\end{minipage}
\hfill
  \begin{minipage}[t]{.45\linewidth}
    \centering
    \begin{tabular}{l @{\hskip 0.6cm} r}
    \toprule
    Level & mIoU \\ \hline
    only TDA   & 43.26  \\ 
    GDA-L1  & 50.72 \\
    GDA-L1,2   & 54.89 \\
    GDA-L1,2,3  & 54.96 \\ \bottomrule
    \end{tabular}
    \caption{Impact of the quality of generated labeled data. }
    \label{tab:generated_quality}
  \end{minipage}
\end{table}

Additionally, we analyze the influence of the quality of generated samples~$\textbf{X}^L_{gda}$ on GDA.
Since there is no ground truth for the generated labeled point clouds, 
we classify them into 3 categories based on visual inspection: (1) L1 - good quality; (2) L2 - few wrong labels or shape artifacts; (3) L3 - low-quality samples with shape distortion or wrong labels.
We rate 100 samples for each category across arbitrary-oriented airplane, car, and lamp classes. The distribution of L1, L2, and L3 samples are 49/46/5, 36/50/14, and 44/49/7 for these categories, respectively.
Examples of three levels are demonstrated in supplementary materials.
We train the segmentation models using combinations of L1, L1+2, and L1+2+3 quality samples, respectively. 
The results listed in Table~\ref{tab:generated_quality} indicate the generated samples of L1 and L2 quality steadily improve the performance of the segmentation model. Although the L3-generated samples are of lower quality, their small proportion ensures they do not adversely affect the training. 

\section{Conclusion}
In this paper, we propose a novel part-aware generative model to produce high-quality labeled point clouds from given segmentation masks.
Leveraging this model, we introduce the first diffusion-based GDA framework for 3D segmentation, which enhances training with diverse 3D variants from limited labeled data and validates pseudo labels using a novel diffusion-based pseudo-label filtering method.
Experiments on three challenging datasets and various segmentation models demonstrate the robustness of GDA and its superiority over TDA and related semi-/self-supervised methods.
As a pioneering work in 3D GDA, this work highlights the potential of GDA in reducing manual labeling efforts and improving segmentation performance.

{
    \small
    \bibliographystyle{ieeenat_fullname}
    \bibliography{main}
}
\appendix
\clearpage
\maketitlesupplementary

\tableofcontents

% \section{Rationale}
% \label{sec:rationale}
% % 
% Having the supplementary compiled together with the main paper means that:
% % 
% \begin{itemize}
% \item The supplementary can back-reference sections of the main paper, for example, we can refer to \cref{sec:intro};
% \item The main paper can forward reference sub-sections within the supplementary explicitly (e.g. referring to a particular experiment); 
% \item When submitted to arXiv, the supplementary will already included at the end of the paper.
% \end{itemize}
% % 
% To split the supplementary pages from the main paper, you can use \href{https://support.apple.com/en-ca/guide/preview/prvw11793/mac#:~:text=Delete%20a%20page%20from%20a,or%20choose%20Edit%20%3E%20Delete).}{Preview (on macOS)}, \href{https://www.adobe.com/acrobat/how-to/delete-pages-from-pdf.html#:~:text=Choose%20%E2%80%9CTools%E2%80%9D%20%3E%20%E2%80%9COrganize,or%20pages%20from%20the%20file.}{Adobe Acrobat} (on all OSs), as well as \href{https://superuser.com/questions/517986/is-it-possible-to-delete-some-pages-of-a-pdf-document}{command line tools}.

\section{More Experimental Results}

\subsection{GDA for Various Segmentation Models on IntrA Dataset}
In the main paper, we test the performance of GDA on IntrA~\citep{yang2020intra} dataset using PointNet~\citep{Qi2016PointNetDL}.
We repeat the experiments using other segmentation models, including PointNet++~\citep{Qi2017PointNetpp} and SPoTr~\citep{Park2023SelfPositioningPT}. 
The results presented in Table~\ref{tab:sup_intra} show that GDA consistently enhances various segmentation models on this real-world medical dataset.

\subsection{Ratio of Labeled Samples in the Training Set}
In the main paper, we conduct experiments with the training sets consisting of 10\% labeled and 90\% unlabeled samples.
Here, we evaluate the impact of different labeled-samples ratios on TDA and GDA in the challenging arbitrary-oriented car class from ShapeNetPart~\citep{Yi2016ASA}. 
As shown in Table~\ref{tab:sup_label_ratio}, GDA outperforms TDA across different label ratios but experiences a more substantial performance drop when only 5\% of the samples are labeled.

\subsection{Impact of mIoU Threshold for Pseudo-label Filtering}
\begin{figure}[t]
    \centering
    \includegraphics[width=0.7\linewidth]{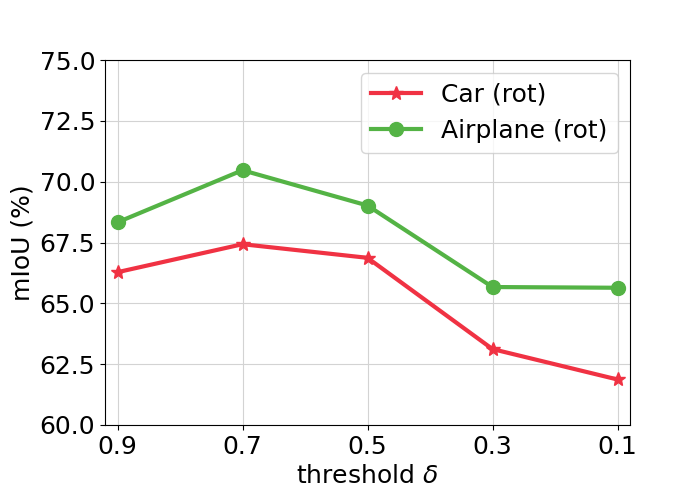}
    \caption{The impact of mIoU threshold $\delta$ for pseudo-label filtering on GDA performance in the airplane and car classes from ShapeNetPart~\citep{Yi2016ASA} (arbitrary orientation).}
    \label{fig:pf_threshold}
\end{figure}

In step 3 of GDA, we filter out the pseudo-labeled samples whose voxelized mIoU between $x^u$ and $\hat{x}^u$ falls below the threshold $\delta$.
To assess the impact of different $\delta$ values on GDA performance, we conduct experiments and present the results in Figure~\ref{fig:pf_threshold}. 
Both the airplane and car categories (arbitrary orientation) achieve the best performance at $\delta = 0.7$.
Hence, we set $\delta$ to 0.7 in all experiments.

\begin{table}[t]
\centering
\renewcommand\arraystretch{1.1}
\begin{tabular}{l c  c }
\toprule
Backbone & Method        & Aneurysm \\ \hline
PointNet & only TDA & 42.29        \\ % \hline
~~~~~\citep{Qi2016PointNetDL} & GDA {\fontsize{8}{8}\selectfont (VG+FP)} & 48.67\\ \hdashline[1pt/2pt]
PointNet++ & only TDA  &  47.15      \\ % \hline
~~~~~\citep{Qi2017PointNetpp} & GDA {\fontsize{8}{8}\selectfont (VG+FP)} & 51.10 \\ \hdashline[1pt/2pt]
\multirow{2}{*}{SPoTr \citep{Park2023SelfPositioningPT}} & only TDA & 48.62          \\ % \hline
& GDA {\fontsize{8}{8}\selectfont (VG+FP)} & \textbf{53.48} \\ \bottomrule
\end{tabular}
\caption{GDA for PointNet~\citep{Qi2016PointNetDL}, PointNet++ \citep{Qi2017PointNetpp}, and SPoTr \citep{Park2023SelfPositioningPT} on IntrA~\citep{yang2020intra} dataset.}
\label{tab:sup_intra}
\end{table}

\begin{table}[t]
    \centering
    \renewcommand\arraystretch{1.1}
    \begin{tabular}{l c c}
    \toprule
       Label Ratio  & TDA & GDA \\ \hline
       5\%  & 27.42 & 55.31 \\
       10\% & 31.09 & 67.43 \\
       20\% & 40.72 & 69.15 \\ \bottomrule
    \end{tabular}
    \caption{The impact of label ratio on TDA and GDA in the car class from ShapeNetPart~\citep{Yi2016ASA} (arbitrary orientation).}
    \label{tab:sup_label_ratio}
\end{table}

\begin{table}[t]
    \centering
    \renewcommand\arraystretch{1.1}
    \begin{tabular}{c c c}
        \toprule
        Part-aware Global Prior & GA in All Layers & mIoU \\ \hline
         \checkmark & & 66.26 \\
          & \checkmark &  65.47 \\
          \checkmark & \checkmark & \textbf{67.43} \\
         \bottomrule
    \end{tabular}
    \caption{Ablation study on the part-aware generative model in the car class from ShapeNetPart~\citep{Yi2016ASA} (arbitrary orientation).}
    \label{tab:supp_ablation}
\end{table}

\subsection{Ablation Study on the Part-aware Generative Model}
Besides integrating segmentation conditioning (SC) modules into local-level modules for point-wise feature prediction given segmentation masks $y$, we additionally extend Lion~\citep{zeng2022lion} in two aspects: 
\textbf{1.}~We incorporate segmentation encoding into both global encoder and global diffusion to provide a more informative global prior.
\textbf{2.}~We integrate global attention (GA) modules into all layers of {p-PVCNN} to enhance intra-part feature encoding, particularly for small parts.
To assess the impact of these modifications, we conduct ablation studies. The results in Table~\ref{tab:supp_ablation} demonstrate that both improvements enhance the part-aware generative model and ultimately benefit downstream segmentation training.

\section{Discussion: The Impact of Label Quality}
\begin{figure}[t]
    \centering
    \includegraphics[width=0.9\linewidth]{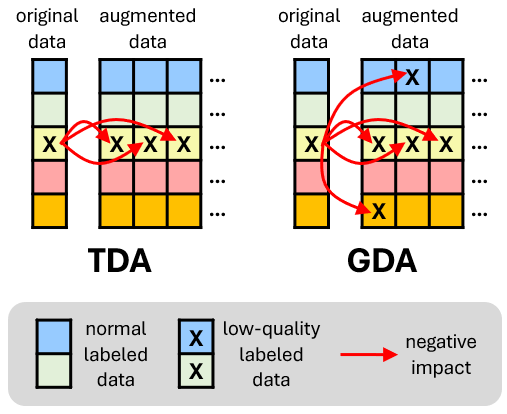}
    \caption{Impact of low-quality hand-labeled samples on TDA and GDA methods. In TDA, low-quality labeled samples mainly affect their own augmented variants. In GDA, the low-quality labeled samples degrade the generative model, which in turn affects the variant generation of other samples.}
    \label{fig:supp_sensitivity}
\end{figure}

In the ablation study of the main paper, we assess the impact of hand-labeled data quality on TDA and GDA performance. Table~6 of the main paper shows that GDA suffers a more significant performance drop than TDA when labeled data quality is not verified, highlighting its sensitivity to labeling accuracy. Figure~\ref{fig:supp_sensitivity} provides further analysis of this phenomenon. In TDA, low-quality labeled samples mainly affect their own augmented variants, such as randomly rotated or flipped versions. However, in GDA, the impact is broader. Poorly labeled samples degrade the generative model, which in turn affects the variant generation of other samples and amplifies the overall negative impact.

\section{More Implementation \& Experiment Details}
\noindent \textbf{Global Encoder $\phi_z$.}
The architecture of the global encoder $\phi_z$ is illustrated in Fig.~\ref{fig:global_latent}~(a).
The point cloud $x \in \mathbb{R}^{n \times 3}$ is firstly concatenated with the segmentation label / zero padding $y \in \mathbb{R}^{n \times c}$, where $n$ and $c$ are respectively the number of points and part types.
The points capture the neighboring features in point-voxel convolutions (PVConv)~\cite{liu2019point}, and they are down-sampled from 2048 to 1024 and eventually to 256 points in set abstraction (SA) layers~\cite{Qi2017PointNetpp}.
The max pooling layer outputs a global intermediate feature, and the multilayer perceptron (MLP) transforms it to the global latent $z_0 \in \mathbb{R}^{d_z}$.
More details about the hyper-parameters of $\phi_z$ are listed in Table~\ref{tab:global_encoder}.
The PVConv~\cite{liu2019point} module mentioned above is illustrated in Fig.~\ref{fig:global_latent}~(c). 
It combines the advantages of point-based and voxel-based methods. The point-based pipeline consists of a linear layer, a group normalization (GN)~\cite{wu2018group}, and a swish activation function. The voxel-based pipeline additionally contains 3D convolution, dropout, and the squeeze-and-excitation (SE)~\cite{hu2018squeeze} layers. 
In the SE layer, the feature is multiplied by an adaptively calibrated channel-wise factor.
The outputs from these two pipelines are merged by summation.

\begin{figure}[t]
    \centering
    \includegraphics[width=0.98\linewidth]{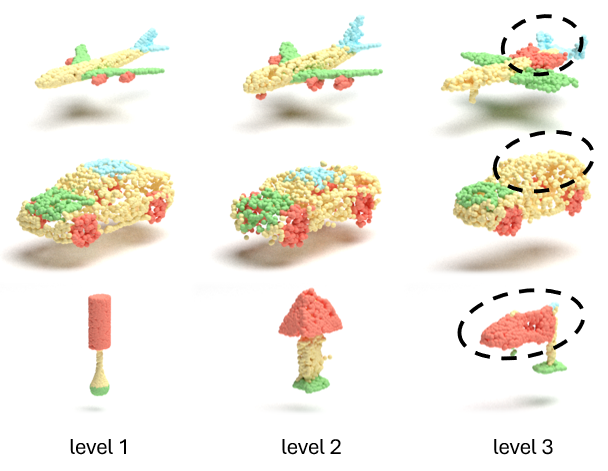}
    \caption{Examples of generated samples from level 1, 2, and 3. The level 1 samples exhibit high-quality shapes and accurate point-wise labels. Although the level 2 samples contain artifacts of jittering points or non-uniformly distributed points, it generally maintains a reasonable shape and segmentation labels. On the level 3 samples, the problematic parts are highlighted in black dashed circles.
    }
    \label{fig:ranking}
\end{figure}

\begin{figure*}[t]
    \centering
    \includegraphics[width=0.8\linewidth]{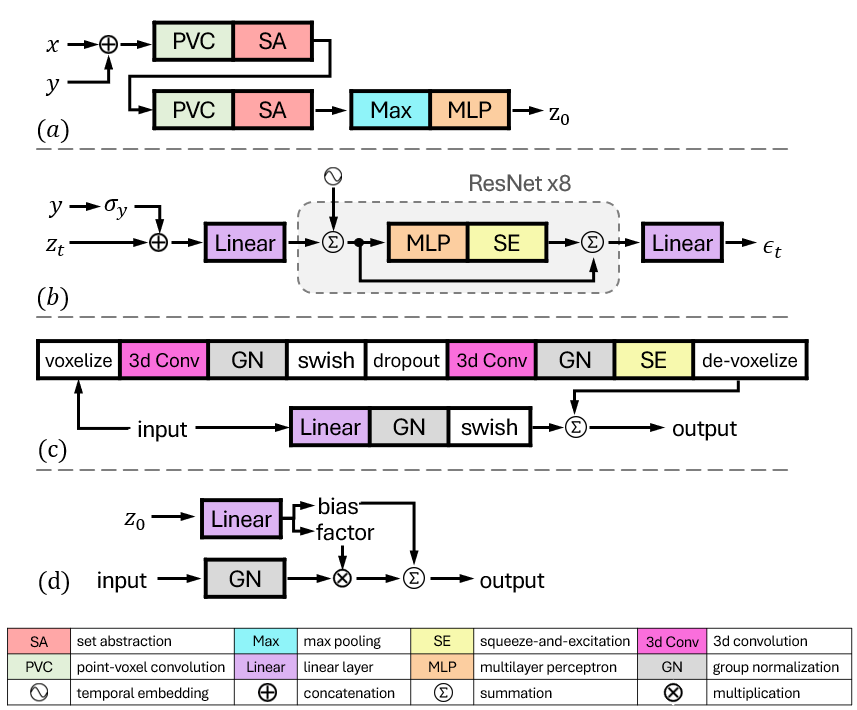}
    \caption{The architecture of (a) the global encoder $\phi_z$, (b) the global diffusion  $\epsilon_z$, (c) the point-voxel convolutional module, and (d) the adaptive group normalization layer.
    }
    \label{fig:global_latent}
\end{figure*}

\noindent \textbf{Global Diffusion $\epsilon_z$.}
The architecture of the global diffusion $\epsilon_z$ is illustrated in Fig.~\ref{fig:global_latent}~(b).
The segmentation label / zero padding $y\in \mathbb{R}^{n \times c}$ is firstly accumulated along the point dimension to obtain $\sigma_y \in \mathbb{R}^c$. For the labeled samples, $\sigma_y$ is a vector of the number of points for each part, while for the unlabeled samples, it is a zero vector.
Then $\sigma_y$ is concatenated the noisy global latent $z_t$ at time step $t$ and fed to a linear layer. 
The core of the global diffusion module is the stacked ResNet~\cite{he2016deep}, 
in which the feature is first fused with temporal embedding by summation and then fused with the output from MLP and SE~\cite{hu2018squeeze} layer.
The output of the stacked ResNet is eventually transformed into the predicted noise through the output linear layer.
More details about the hyper-parameters of $\epsilon_z$ are listed in Table~\ref{tab:global_diffusion}.

\noindent \textbf{Point-level Components.}
In our generative model, the point-level encoder $\phi_h$ and decoder $\xi_h$, and the point-level diffusion $\epsilon_h$ adopt a similar 4-layer part-aware PVCNN (p-PVCNN), whose architecture is illustrated in Figure 2 (b) of the main paper. 
PVConv is also used in p-PVCNN, but the GN~\cite{wu2018group} layer is replaced by an adaptive group normalization~\cite{zeng2022lion} layer for the conditioning on the global latent $z_0$.
The pipeline of the adaptive GN is shown in Fig.~\ref{fig:global_latent} (d). 
More details about the hyper-parameters of $\phi_h$, $\xi_h$, and $\epsilon_h$ are respectively listed in Table~\ref{tab:point_encoder}, \ref{tab:point_decoder}, and \ref{tab:point_diffusion}.

\noindent \textbf{More Experimental Details.}
For the inference of the generative model on an RTX 3090 GPU, each generated sample takes 0.62 seconds on average.
In the training of segmentation models, for the augmentation methods with fewer training samples, e.g. method~2 (w/o aug.), we duplicate samples to maintain a consistent iteration count per training epoch.

\section{More Visualizations}

\subsection{Visualization of Generated Samples of Different Quality Level}
In ablation studies, we classify the generated labeled samples into 3 categories based on visual inspection: (1) Level 1 - good quality; (2) Level 2 - few wrong labels or shape artifacts; (3) Level 3 - low-quality samples with shape distortion or wrong labels.
We demonstrate examples of these three levels from airplane, car, and lamp classes in Figure~\ref{fig:ranking}.

\subsection{Visualization of Generated Variants}
\begin{figure}[t]
    \centering
    \includegraphics[width=0.98\linewidth]{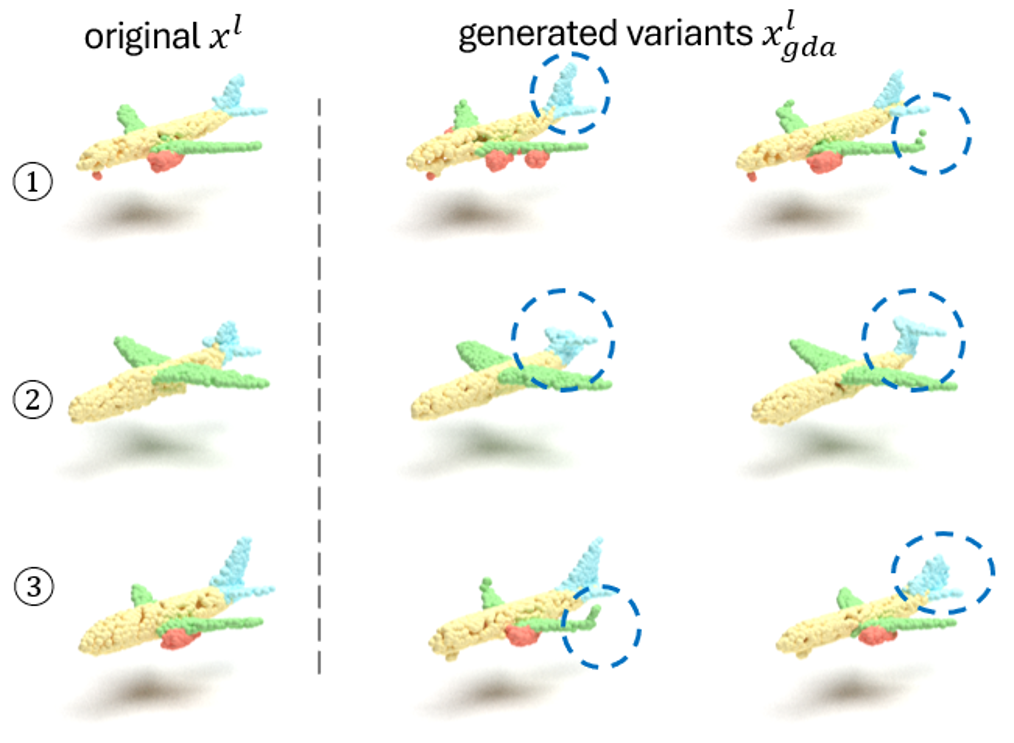}
    \caption{Generated airplane variants. The shape variations are highlighted in blue dashed circles.}
    \label{fig:supp_vg_airplane}
\end{figure}

\begin{figure}[t]
    \centering
    \includegraphics[width=0.98\linewidth]{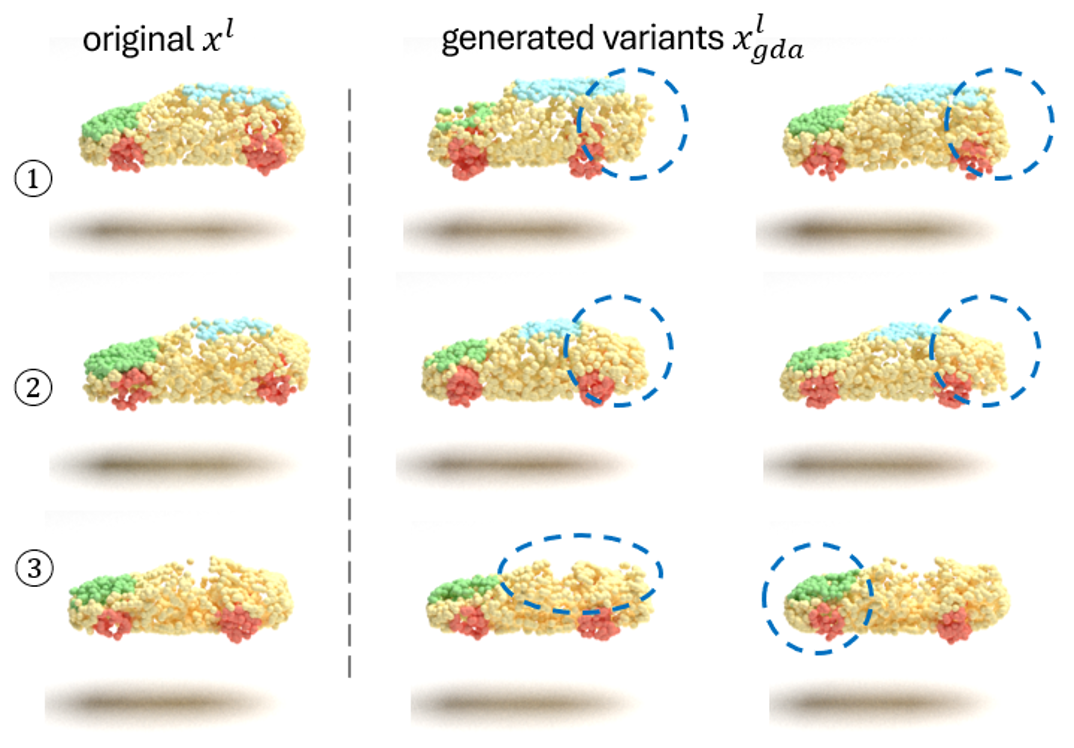}
    \caption{Generated car variants. The shape variations are highlighted in blue dashed circles.}
    \label{fig:supp_vg_car}
\end{figure}

\begin{figure}[t]
    \centering
    \includegraphics[width=0.98\linewidth]{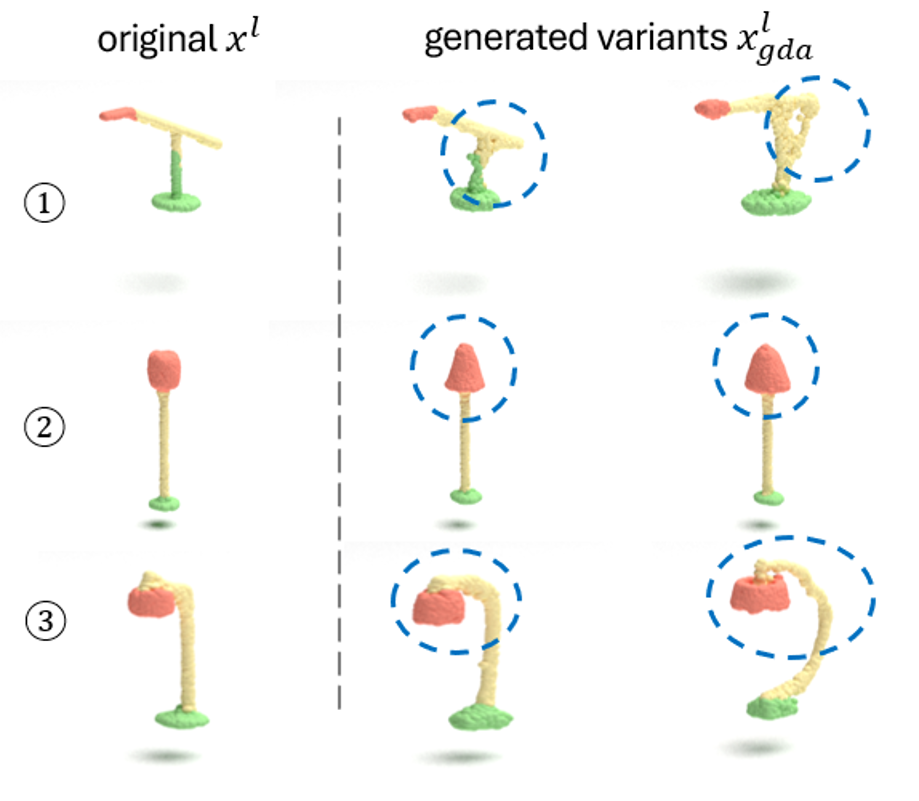}
    \caption{Generated lamp variants. The shape variations are highlighted in blue dashed circles.}
    \label{fig:supp_vg_lamp}
\end{figure}

\begin{figure}[t]
    \centering
    \includegraphics[width=0.98\linewidth]{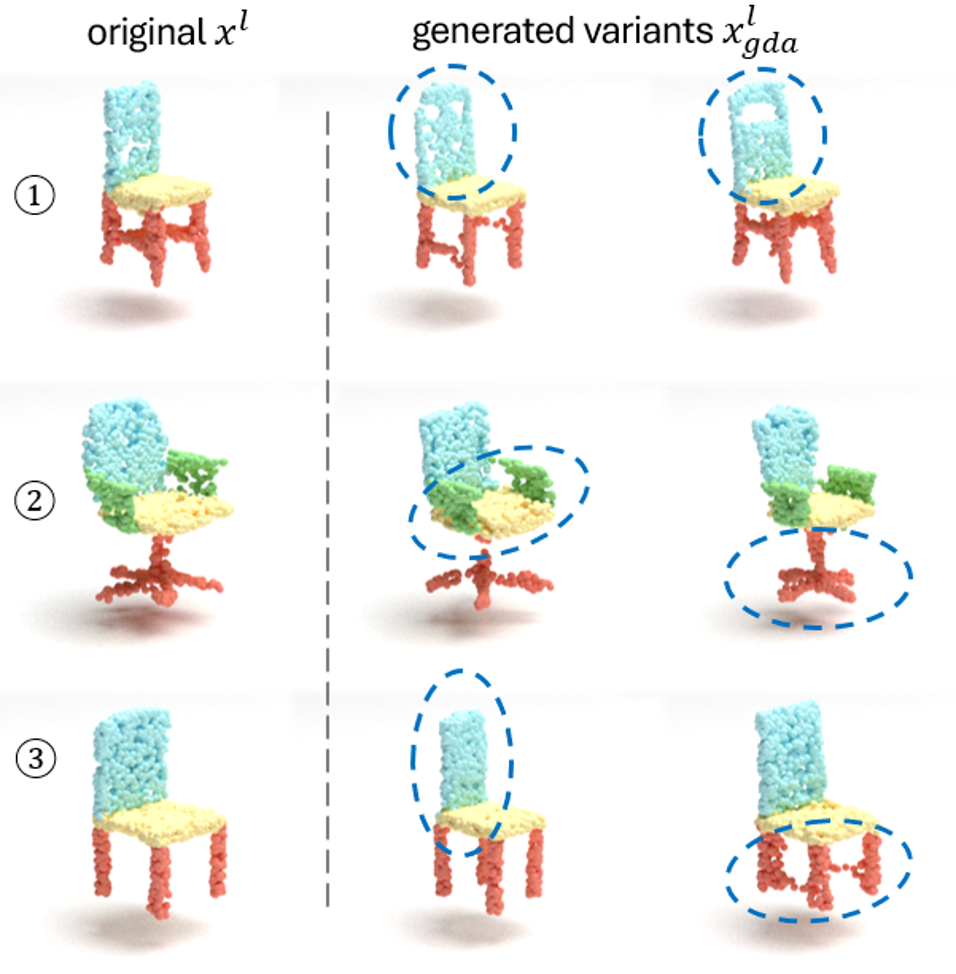}
    \caption{Generated chair variants. The shape variations are highlighted in blue dashed circles.}
    \label{fig:supp_vg_chair}
\end{figure}

\begin{figure}[t]
    \centering
    \includegraphics[width=0.98\linewidth]{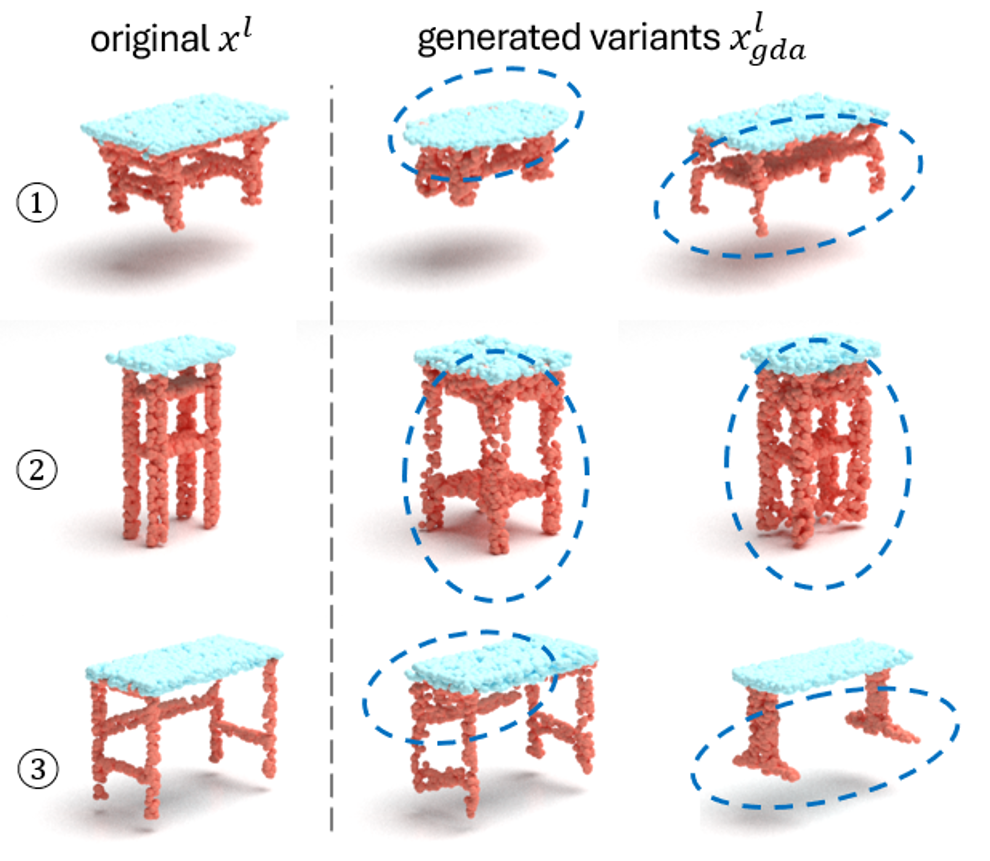}
    \caption{Generated table variants. The shape variations are highlighted in blue dashed circles.}
    \label{fig:supp_vg_table}
\end{figure}

We demonstrate more generated variants alongside the original labeled samples for the airplane, car, lamp, chair, and table categories in Figure~\ref{fig:supp_vg_airplane}, \ref{fig:supp_vg_car}, \ref{fig:supp_vg_lamp}, \ref{fig:supp_vg_chair}, and \ref{fig:supp_vg_table}, respectively.

\subsection{Visualization of Pseudo-label Filtering}
\begin{figure}[t]
    \centering
    \includegraphics[width=0.75\linewidth]{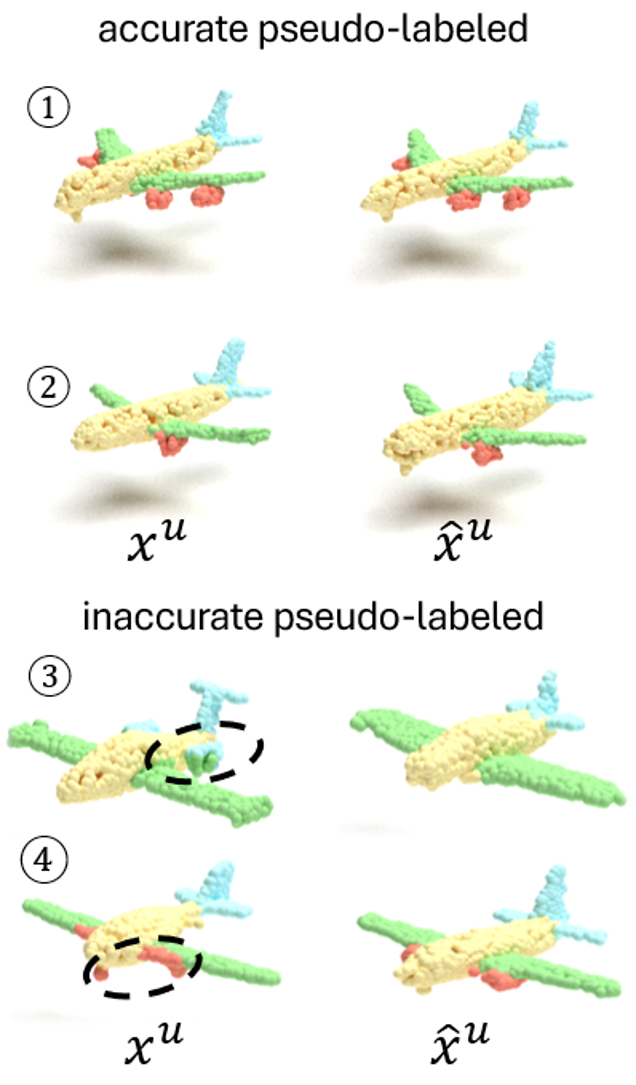}
    \caption{Pseudo-labeled airplanes $x^u$ and the reconstructed samples $\hat{x}^u$ after the conditional diffuse-denoise process. Samples $x^u$ with accurate pseudo labels are closely reconstructed, whereas those with inaccurate pseudo labels undergo substantial shape deviation, especially in the misclassified parts, highlighted by \dotuline{black dashed circles}. Note: The engine (red) of \ding{174} is misclassified as the tail (blue), while the front part of wing (green) of \ding{175} is misclassified as the engine (red).}
    \label{fig:supp_fp_airplane}
\end{figure}

\begin{figure}[t]
    \centering
    \includegraphics[width=0.75\linewidth]{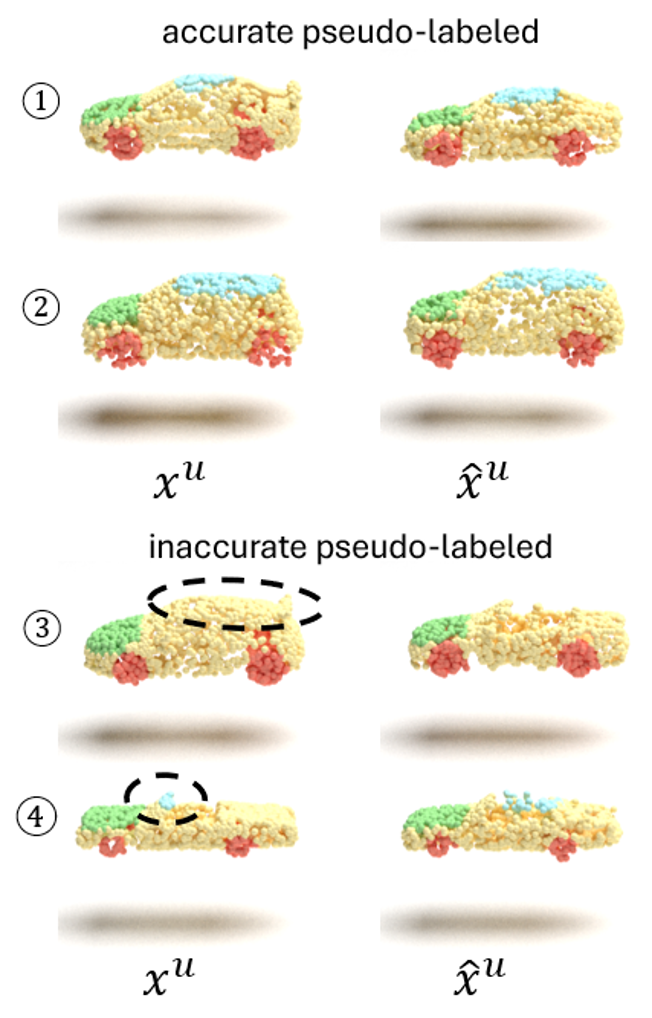}
    \caption{Pseudo-labeled cars $x^u$ and their reconstructed samples $\hat{x}^u$ after the conditional diffuse-denoise process. The misclassified parts are highlighted by \dotuline{black dashed circles}. Note: In \ding{174}, the roof (blue) is omitted, causing the reconstructed car to transform into a convertible to accommodate this misclassification.
In \ding{175}, the tip of the front glass of the convertible is misclassified as a roof (blue), leading to the repositioning of these points backward to ``form" a roof.}
    \label{fig:supp_fp_car}
\end{figure}

\begin{figure}[t]
    \centering
    \includegraphics[width=0.75\linewidth]{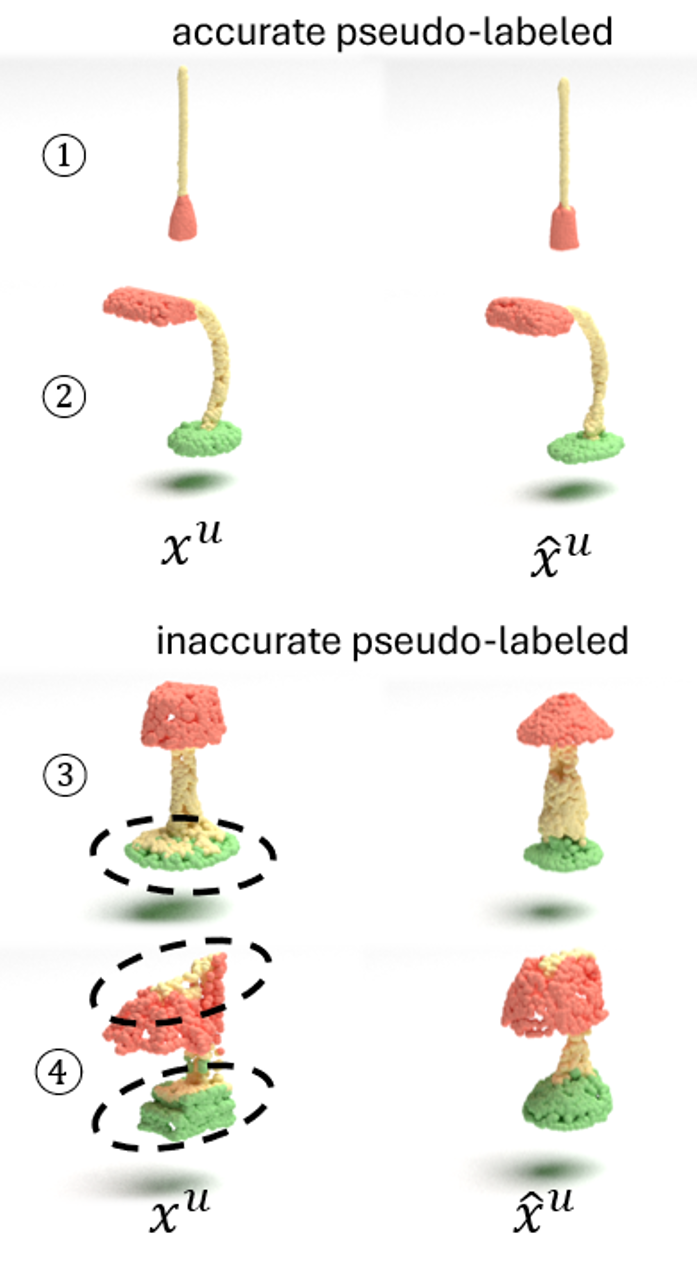}
    \caption{Pseudo-labeled lamps $x^u$ and their reconstructed samples $\hat{x}^u$ after the conditional diffuse-denoise process. Samples $x^u$ with accurate pseudo labels are closely reconstructed, whereas those with inaccurate pseudo labels undergo substantial shape deviation, especially in the misclassified parts, highlighted by \dotuline{black dashed circles}. Note: In \ding{174} and \ding{175}, the upper part of the lamp bases (green) is misclassified as the pole part (yellow).}
    \label{fig:supp_fp_lamp}
\end{figure}

\begin{figure}[t]
    \centering
    \includegraphics[width=0.75\linewidth]{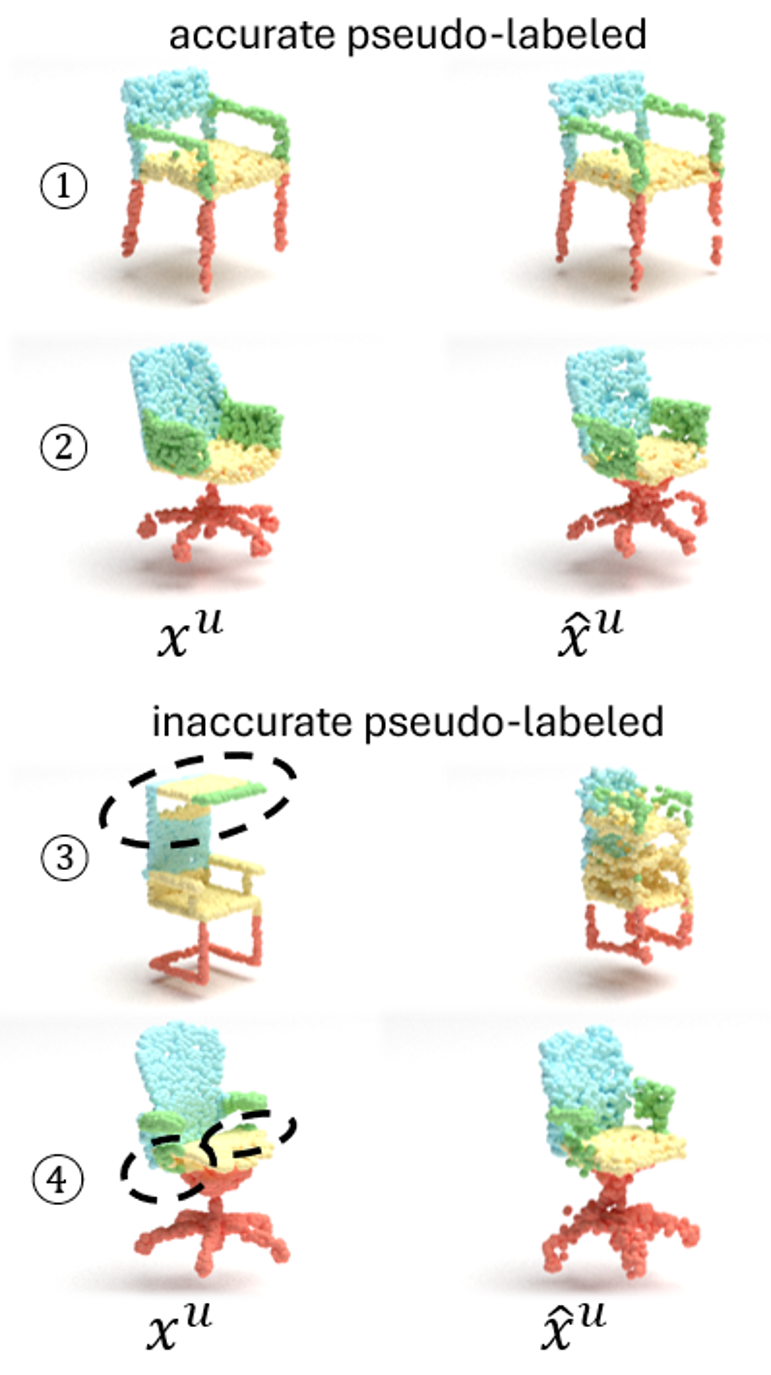}
    \caption{Pseudo-labeled chairs $x^u$ and their reconstructed samples $\hat{x}^u$ after the conditional diffuse-denoise process. The misclassified parts are highlighted by \dotuline{black dashed circles}. Note: The top of \ding{174} is misclassified as the base (yellow) and arm (green) parts, resulting in the repositioning of these points downward and an extremely noisy reconstructed sample. For \ding{175}, the boundary of the base (yellow) part is misclassified as the arm (green) part.}
    \label{fig:supp_fp_chair}
\end{figure}

\begin{figure}[t]
    \centering
    \includegraphics[width=0.75\linewidth]{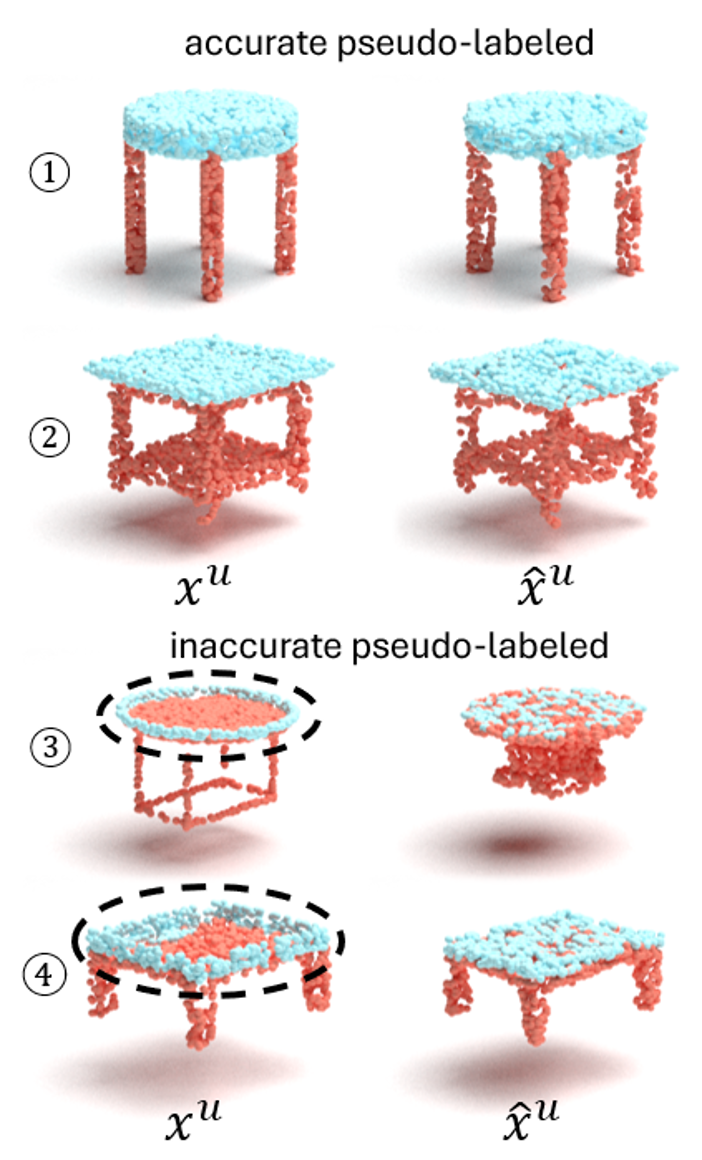}
    \caption{Pseudo-labeled tables $x^u$ and their reconstructed samples $\hat{x}^u$ after the conditional diffuse-denoise process. The misclassified parts are highlighted by \dotuline{black dashed circles}. Note: The central parts of the tabletop (blue) of \ding{174} and \ding{175} are misclassified as the base (red) part, resulting in the ``sparse" tabletop in the reconstructed samples.}
    \label{fig:supp_fp_table}
\end{figure}

We demonstrate more examples of reconstructed samples with accurate and inaccurate pseudo labels after the conditional diffuse-denoise process in Figure~\ref{fig:supp_fp_airplane}, \ref{fig:supp_fp_car}, \ref{fig:supp_fp_lamp}, \ref{fig:supp_fp_chair}, and \ref{fig:supp_fp_table}, respectively.

\subsection{Qualitative Results of Segmentation}
\begin{figure}[t]
    \centering
    \includegraphics[width=0.9\linewidth]{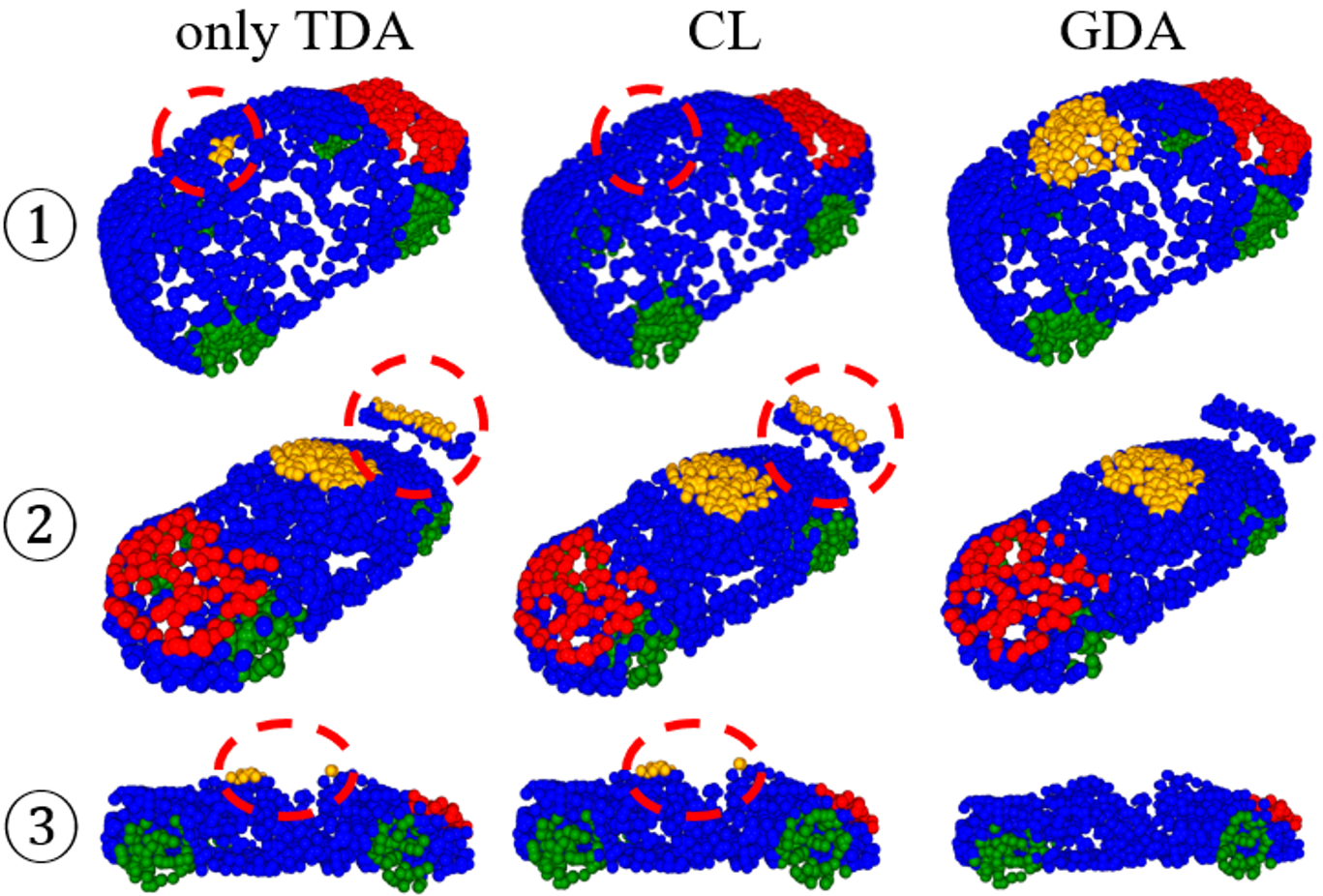}
    \caption{Segmentation predictions on three cars from ShapeNetPart~\cite{Yi2016ASA} using PointNet~\cite{Qi2016PointNetDL} trained with TDA (\textit{left}), CL (\textit{Middle}), and GDA (\textit{Right}).
    The model trained with GDA outperforms the others. Areas of misprediction are highlighted in dashed circles.
    % using different approaches. Note: wrong predicted areas are highlighted in the red circles. \textit{Left:} traditional data augmentation (TDA). \textit{Middle:} semi-supervised method based on contrastive learning (CL). \textit{Right:} generative data augmentation (GDA). 
    }
    \label{fig:seg_car}
\end{figure}

\begin{figure}[t]
    \centering
    \includegraphics[width=0.9\linewidth]{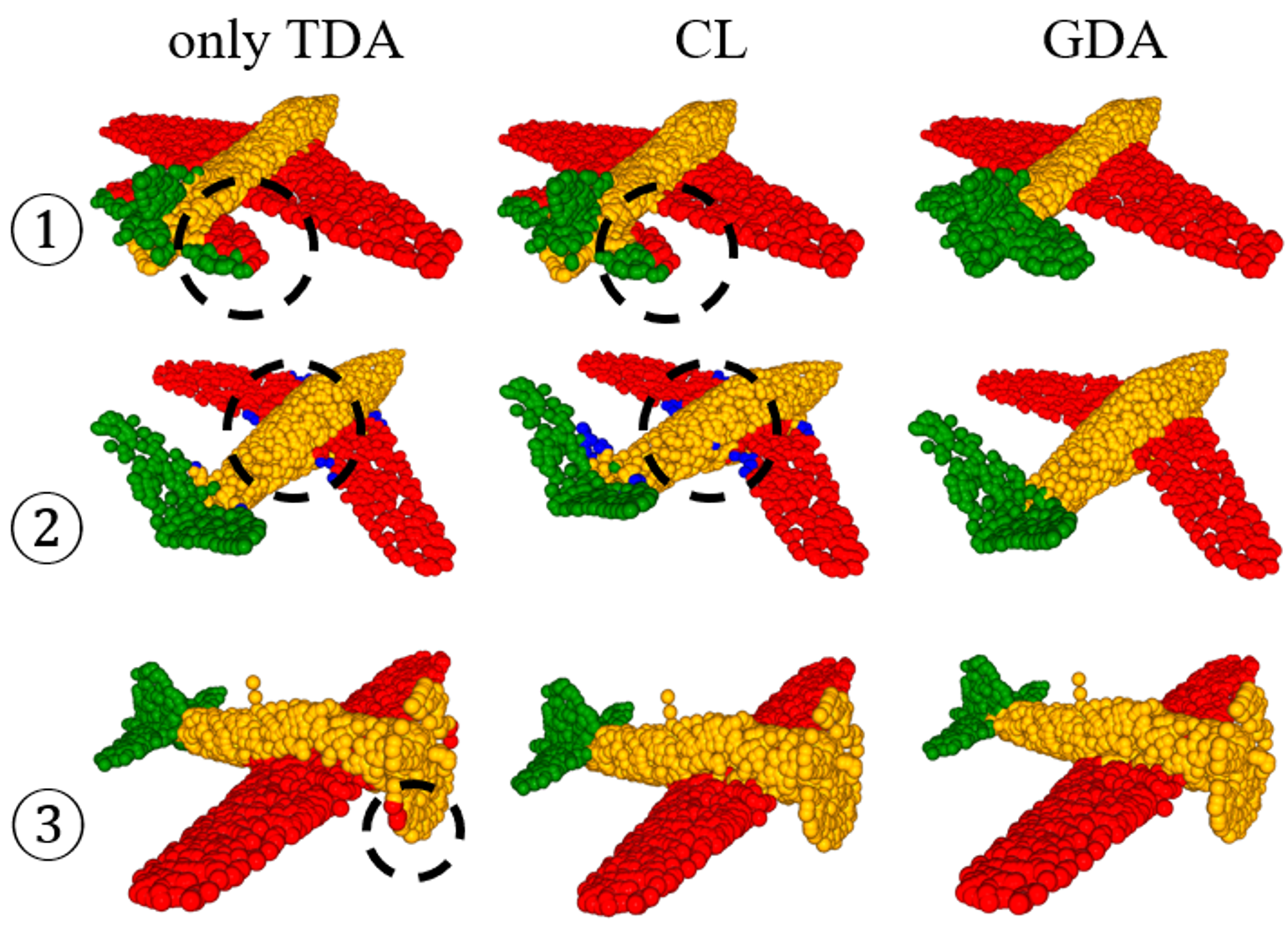}
    \caption{
    Segmentation predictions on three airplanes from ShapeNetPart~\cite{Yi2016ASA} using PointNet~\cite{Qi2016PointNetDL} trained with TDA (\textit{left}), CL (\textit{Middle}), and GDA (\textit{Right}).
    The model trained with GDA outperforms the others. Areas of misprediction are highlighted in dashed circles.
    % Segmentation results on three airplanes (ShapeNet) from PointNet trained using different approaches. Note: wrong predicted areas are highlighted in the black circles. \textit{Left:} traditional data augmentation (TDA). \textit{Middle:} semi-supervised method based on contrastive learning (CL). \textit{Right:} generative data augmentation (GDA). 
    }
    \label{fig:seg_airplane}
\end{figure}

\begin{figure}[t]
    \centering
    \includegraphics[width=0.98\linewidth]{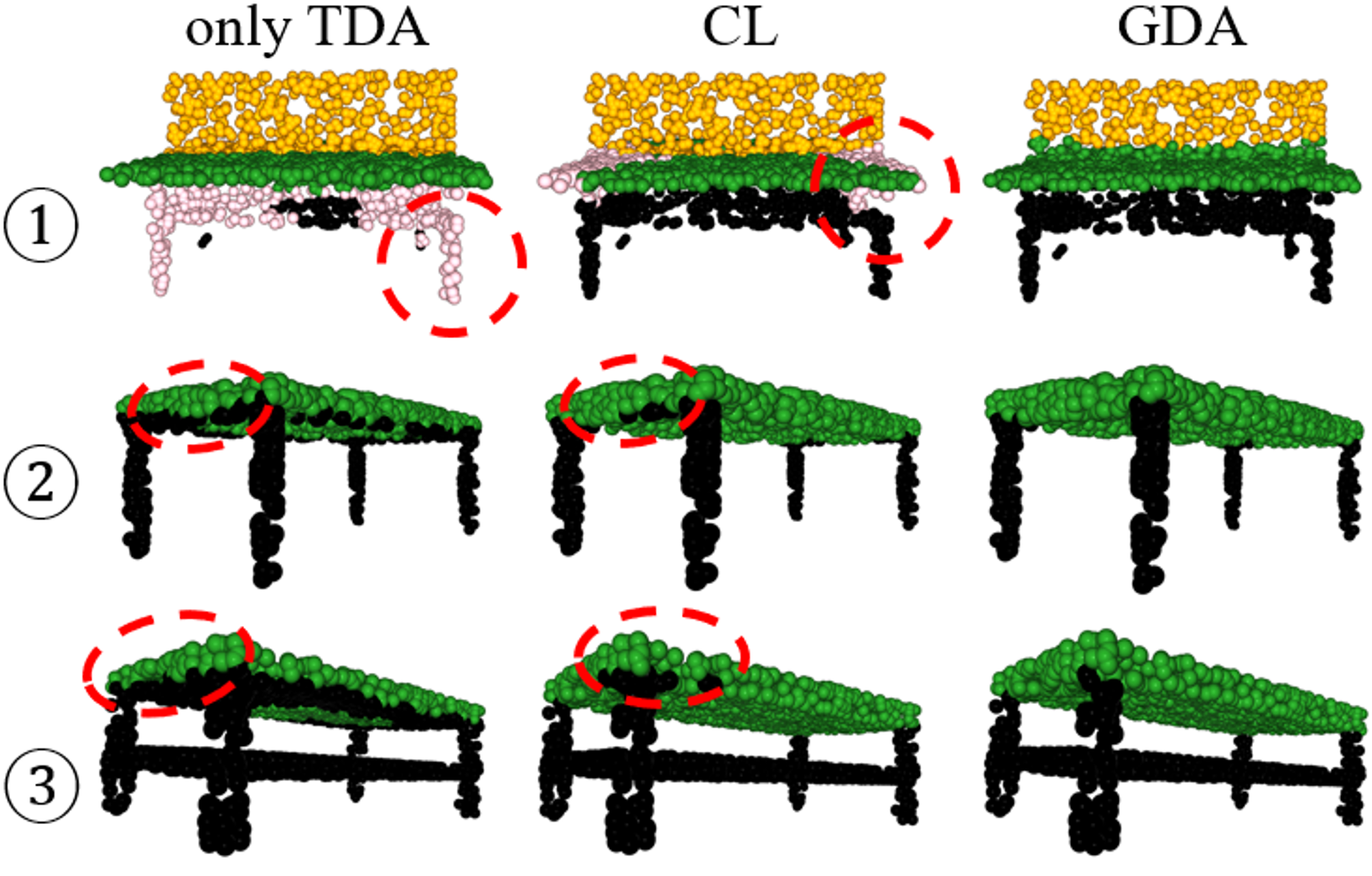}
    \caption{
    Segmentation predictions on three tables from PartNet~\cite{mo2019partnet} using PointNet~\cite{Qi2016PointNetDL} trained with TDA (\textit{left}), CL (\textit{Middle}), and GDA (\textit{Right}).
    The model trained with GDA outperforms the others. Areas of misprediction are highlighted in dashed circles.
    % Segmentation results on three tables (PartNet) from PointNet trained using different approaches. Note: wrong predicted areas are highlighted in the red circles. \textit{Left:} traditional data augmentation (TDA). \textit{Middle:} semi-supervised method based on contrastive learning (CL). \textit{Right:} generative data augmentation (GDA). 
    }
    \label{fig:seg_table}
\end{figure}

\begin{figure}[t]
    \centering
    \includegraphics[width=0.9\linewidth]{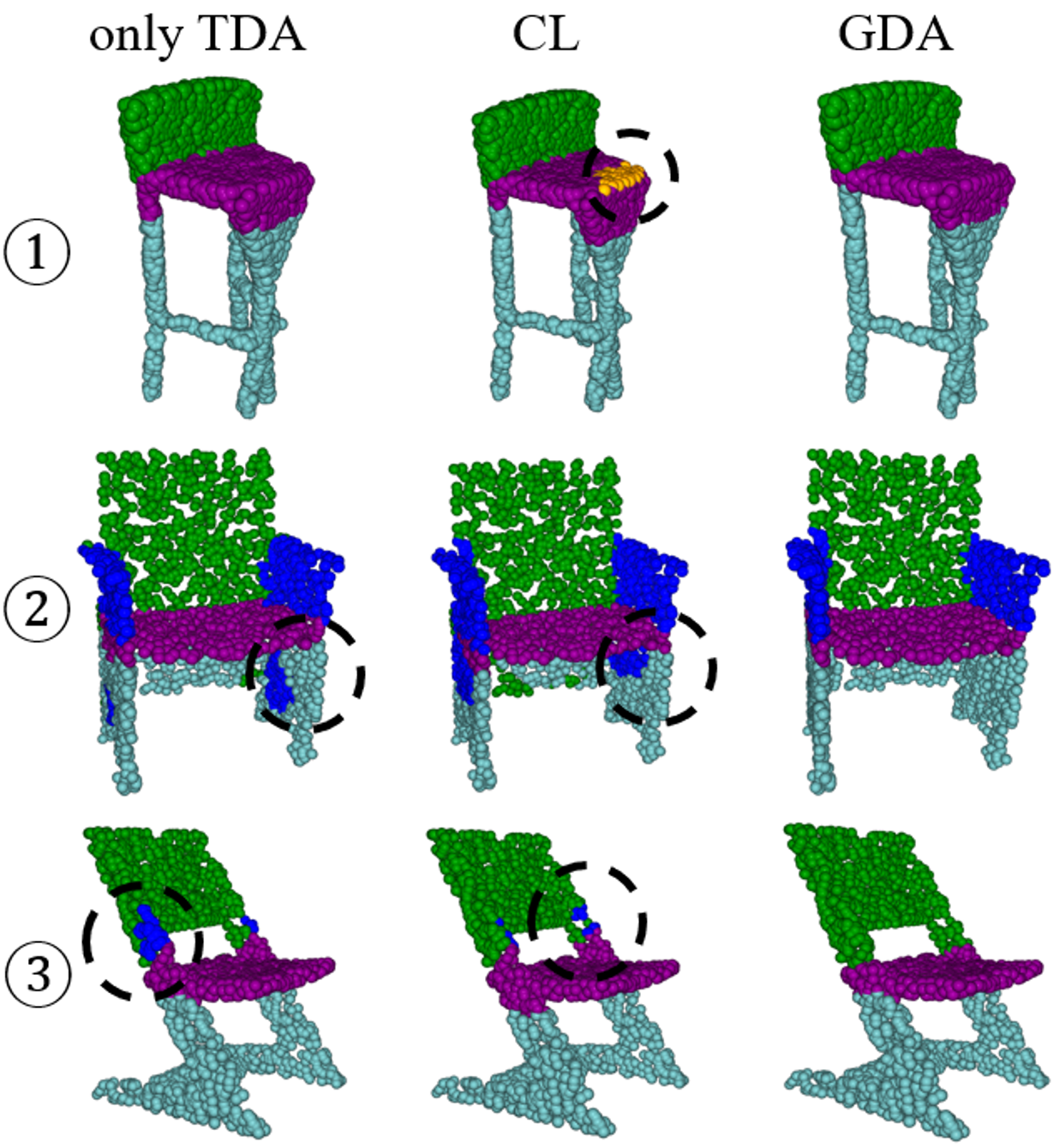}
    \caption{
    Segmentation predictions on three chairs from PartNet~\cite{Yi2016ASA} using PointNet~\cite{Qi2016PointNetDL} trained with TDA (\textit{left}), CL (\textit{Middle}), and GDA (\textit{Right}).
    The model trained with GDA outperforms the others. Areas of misprediction are highlighted in dashed circles.
    % Segmentation results on three chairs (PartNet) from PointNet trained using different approaches. Note: wrong predicted areas are highlighted in the black circles. \textit{Left:} traditional data augmentation (TDA). \textit{Middle:} semi-supervised method based on contrastive learning (CL). \textit{Right:} generative data augmentation (GDA). 
    }
    \label{fig:seg_chair}
\end{figure}

Additionally, this supplementary material provides some qualitative results of the experiments on ShapeNetPart~\cite{Yi2016ASA} and PartNet~\cite{mo2019partnet}. 
We show the part segmentation results from PointNet~\cite{Qi2016PointNetDL} trained with different approaches: 
\begin{enumerate}
    \item only using traditional data augmentation (TDA),
    \item using a semi-supervised method based on contrastive learning (CL)~\cite{jiang2021guided},
    \item using generative data augmentation (GDA) based on variant generation and filtered pseudo labels.
\end{enumerate}
The segmentation results on cars and airplanes from ShapeNetPart~\cite{Yi2016ASA} are demonstrated in Fig.~\ref{fig:seg_car} and Fig.~\ref{fig:seg_airplane} respectively.
The segmentation results on tables and chairs from PartNet~\cite{mo2019partnet} are demonstrated in Fig.~\ref{fig:seg_table} and Fig.~\ref{fig:seg_chair} respectively.
The segmentation neural network trained with GDA significantly outperforms the others.
For example, \ding{173} in Figure~\ref{fig:seg_car} shows rear spoilers of the car body are mis-predicted as roof by models trained with TDA or CL, while the model trained with GDA predicts them correctly.

\begin{table*}[t]
\centering
\renewcommand\arraystretch{1.2}
\begin{tabular}{@{}c|l|c|c@{}}
\toprule
Input & \multicolumn{3}{l}{point clouds ($2048 \times 3$), segmentation labels ($2048 \times c$)} \\ \hline
% \multicolumn{4}{l}{Input: segmentation labels ($2048 \times c$)} \\ \hline
Output & \multicolumn{3}{l}{global latent ($1 \times 128$)} \\ \hline \hline
\multicolumn{2}{l|}{} & Layer 1 & Layer 2 \\ \hline
\multirow{3}{*}{PVConv} & layers & 2 & 1 \\
& hidden dimensions & 32 & 32 \\
& voxel grid size & 32 & 16 \\ \hline
\multirow{5}{*}{SA} & grouper center & 1024 & 256 \\
& grouper radius & 0.1 & 0.2 \\
& grouper neighbors & 32 & 32 \\
& MLP layers & 2 & 2 \\
& MLP output dimensions & 32, 32 & 32, 64 \\ \hline \hline
\multirow{2}{*}{Output layer} & MLP layers & \multicolumn{2}{c}{2} \\
& MLP output dimensions & \multicolumn{2}{c}{128, 128} \\
\bottomrule
\end{tabular}
\caption{Hyper-parameters of the global encoder $\phi_z$.}
\label{tab:global_encoder}
\end{table*}

\begin{table*}[t]
\centering
\renewcommand\arraystretch{1.2}
\begin{tabular}{@{}c|l|c@{}}
\toprule
Input & \multicolumn{2}{l}{\makecell[l]{global latent ($1 \times 128$), diffusion time step $t$ \\ segmentation labels ($2048 \times c$) } }\\ \hline
Output & \multicolumn{2}{l}{predicted noise on global latent ($1 \times 128$)} \\ \hline \hline
Input linear layer & output dimension & 2048 \\ \hline
\multirow{3}{*}{\makecell[c]{Time embedding\\layer}} & sinusoidal embedding dimension & 128 \\
& MLP layers & 2 \\
& MLP output dimensions & 512, 2048 \\ \hline
\multirow{4}{*}{Stacked ResNet} & MLP layers & 2 \\
& MLP output dimensions & 2048, 2048 \\
& SE MLP layers & 2 \\
& SE MLP output dimensions & 256, 2048 \\ \hline
Output linear layer & output dimension & 128 \\
\bottomrule
\end{tabular}
\caption{Hyper-parameters of the global diffusion $\epsilon_z$.}
\label{tab:global_diffusion}
\end{table*}

\begin{table*}[t]
\centering
\renewcommand\arraystretch{1.2}
\begin{tabular}{@{}c|l|c|c|c|c@{}}
\toprule
% \multicolumn{6}{l}{Input: point clouds ($2048 \times 3$)} \\
% \multicolumn{6}{l}{Input: segmentation labels ($2048 \times c$)} \\
% \multicolumn{6}{l}{Input: global latent ($1 \times 128$)} \\ \hline
Input & \multicolumn{5}{l}{\makecell[l]{point clouds ($2048\times3$), segmentation labels ($2048 \times c$), \\global latent ($1 \times 128$)}} \\ \hline
% \multicolumn{6}{l}{Input: segmentation labels ($2048 \times c$)} \\
% \multicolumn{6}{l}{Input: global latent ($1 \times 128$)} \\ \hline
Output & \multicolumn{5}{l}{point-level latent ($2048 \times 4$)} \\ \hline \hline 
\multicolumn{2}{l|}{} & Layer 1 & Layer 2 & Layer 3 & Layer 4  \\ \hline
\multirow{3}{*}{PVConv} & layers & 2 & 1 & 1 & - \\
& hidden dimensions & 32 & 64 & 128 & - \\
& voxel grid size & 32 & 16 & 8 & - \\ \hline
\multirow{5}{*}{SA} & grouper center & 1024 & 256 & 64 & 16 \\
& grouper radius & 0.1 & 0.2 & 0.4 & 0.8 \\
& grouper neighbors & 32 & 32 & 32 & 32 \\
& MLP layers & 2 & 2 & 2 & 3 \\
& MLP output dimensions & 32, 32 & 64, 128 & 128, 256 & 128, 128, 128 \\ \hline
\multirow{2}{*}{GA} & hidden dimensions & 32+c & 128+c & 256+c & 128+c \\
& attention heads & 8 & 8 & 8 & 8 \\ \hline \hline
\multirow{2}{*}{FP} & MLP layers & 3 & 2 & 2 & 2 \\
& MLP output dimensions & 128, 128, 64 & 128, 128 & 128, 128 & 128, 128 \\ \hline
\multirow{3}{*}{PVConv} & layers & 2 & 2 & 3 & 3 \\
& hidden dimensions & 64 & 128 & 128 & 128 \\
& voxel grid size & 32 & 16 & 8 & 8 \\
\bottomrule
\end{tabular}
\caption{Hyper-parameters of the point-level encoder $\phi_h$. Note: layer 1 refers to the shallowest layer and layer 4 refers to the deepest layer.}
\label{tab:point_encoder}
\end{table*}

\begin{table*}[t]
\centering
\renewcommand\arraystretch{1.1}
\begin{tabular}{@{}c|l|c|c|c|c@{}}
\toprule
Input & \multicolumn{5}{l}{\makecell[l]{point-level latent ($2048 \times 4$), segmentation labels ($2048 \times c$), \\ global latent ($1 \times 128$) } }\\ \hline
Output & \multicolumn{5}{l}{point cloud ($2048 \times 3$)} \\ \hline \hline 
\multicolumn{2}{l|}{} & Layer 1 & Layer 2 & Layer 3 & Layer 4  \\ \hline
\multirow{3}{*}{PVConv} & layers & 2 & 1 & 1 & - \\
& hidden dimensions & 32 & 64 & 128 & - \\
& voxel grid size & 32 & 16 & 8 & - \\ \hline
\multirow{5}{*}{SA} & grouper center & 1024 & 256 & 64 & 16 \\
& grouper radius & 0.1 & 0.2 & 0.4 & 0.8 \\
& grouper neighbors & 32 & 32 & 32 & 32 \\
& MLP layers & 2 & 2 & 2 & 3 \\
& MLP output dimensions & 32, 64 & 64, 128 & 128, 256 & 128, 128, 128 \\ \hline
\multirow{2}{*}{GA} & hidden dimensions & 64+c & 128+c & 256+c & 128+c \\
& attention heads & 8 & 8 & 8 & 8 \\ \hline \hline
\multirow{2}{*}{FP} & MLP layers & 3 & 2 & 2 & 2 \\
& MLP output dimensions & 128, 128, 64 & 128, 128 & 128, 128 & 128, 128 \\ \hline
\multirow{3}{*}{PVConv} & layers & 2 & 2 & 3 & 3 \\
& hidden dimensions & 64 & 128 & 128 & 128 \\
& voxel grid size & 32 & 16 & 8 & 8 \\ \hline \hline
\multirow{2}{*}{Output layer} & MLP layers & \multicolumn{4}{c}{2} \\
& MLP output dimensions & \multicolumn{4}{c}{128, 3} \\
\bottomrule
\end{tabular}
\caption{Hyper-parameters of the point-level decoder $\xi_h$. Note: layer 1 refers to the shallowest layer and layer 4 refers to the deepest layer.}
\label{tab:point_decoder}
\end{table*}

\begin{table*}[t]
\centering
\renewcommand\arraystretch{1.1}
\begin{tabular}{@{}c|l|c|c|c|c@{}}
\toprule
Input & \multicolumn{5}{l}{\makecell[l]{point-level latent ($2048 \times 4$), diffusion time step $t$ \\ segmentation labels ($2048 \times c$), global latent ($1 \times 128$) } }\\ \hline
Output & \multicolumn{5}{l}{predicted noise on point-level latent ($2048 \times 4$)} \\ \hline \hline 
\multirow{3}{*}{\makecell[c]{Time\\embedding}} & sinusoidal dimensions & \multicolumn{4}{c}{64} \\
& MLP layers & \multicolumn{4}{c}{2} \\
& MLP output dimensions & \multicolumn{4}{c}{64, 64} \\ \hline \hline
\multicolumn{2}{l|}{} & Layer 1 & Layer 2 & Layer 3 & Layer 4  \\ \hline
\multirow{3}{*}{PVConv} & layers & 2 & 1 & 1 & - \\
& hidden dimensions & 32 & 64 & 128 & - \\
& voxel grid size & 32 & 16 & 8 & - \\ \hline
\multirow{5}{*}{SA} & grouper center & 1024 & 256 & 64 & 16 \\
& grouper radius & 0.1 & 0.2 & 0.4 & 0.8 \\
& grouper neighbors & 32 & 32 & 32 & 32 \\
& MLP layers & 2 & 2 & 2 & 3 \\
& MLP output dimensions & 32, 64 & 64, 128 & 128, 256 & 128, 128, 128 \\ \hline
\multirow{2}{*}{GA} & hidden dimensions & 64+c & 128+c & 256+c & 128+c \\
& attention heads & 8 & 8 & 8 & 8 \\ \hline \hline
\multirow{2}{*}{FP} & MLP layers & 3 & 2 & 2 & 2 \\
& MLP output dimensions & 128, 128, 64 & 128, 128 & 128, 128 & 128, 128 \\ \hline
\multirow{3}{*}{PVConv} & layers & 2 & 2 & 3 & 3 \\
& hidden dimensions & 64 & 128 & 128 & 128 \\
& voxel grid size & 32 & 16 & 8 & 8 \\ \hline \hline
\multirow{2}{*}{Output layer} & MLP layers & \multicolumn{4}{c}{2} \\
& MLP output dimensions & \multicolumn{4}{c}{128, 4} \\
\bottomrule
\end{tabular}
\caption{Hyper-parameters of the point-level diffusion $\epsilon_h$. Note: layer 1 refers to the shallowest layer and layer 4 refers to the deepest layer.}
\label{tab:point_diffusion}
\end{table*}

\clearpage


\end{document}